\documentclass[twoside,11pt]{article}
\usepackage{graphicx}
\usepackage{float}
\usepackage{caption, subcaption}
\usepackage[ruled, linesnumbered]{algorithm2e}
\usepackage{algorithmicx}
\usepackage{algpseudocode}
\usepackage{booktabs}
\usepackage{multirow}
\usepackage{longtable}
\usepackage{colortbl}
\usepackage{array}
\usepackage{xcolor}
\usepackage{setspace}
\usepackage{amsfonts}
\usepackage{amsmath,amssymb}
\usepackage{amsthm}
\usepackage{bm}
\usepackage{url}
\usepackage{enumitem}
\usepackage[a4paper,left=31.7mm,right=31.7mm,top=25.4mm,bottom=25.4mm]{geometry}

\SetKwComment{Comment}{(}{)}
\renewcommand{\KwData}{\textbf{Input: }}
\renewcommand{\KwResult}{\textbf{Output: }}

\title{\textbf{Clustering with Uniformity- and Neighbor-Based Random Geometric Graphs}}
\author{Rui Shi, Elvan Ceyhan, and Nedret Billor \\Department of Mathematics and Statistics\\Auburn University}

\begin{document}
\maketitle

\begin{abstract}
\noindent
We propose a graph-based clustering method based on Cluster Catch Digraphs (CCDs) that extends their applicability to moderate-dimensional data settings. Existing CCD variants, such as RK-CCDs, rely on spatial randomness tests based on Ripley's $K$ function, which exhibit performance degradation as dimensionality increases. To address this limitation, we introduce a nearest-neighbor-distance (NND) based Monte Carlo spatial randomness test (MC-SRT) for determining covering radii, resulting in the proposed Uniformity- and Neighbor-based CCDs (UN-CCDs). The proposed method is designed for datasets of moderate size and dimension, particularly in settings with complex cluster geometry and uniformly distributed background noise. Through Monte Carlo simulations and experiments on benchmark datasets, we show that UN-CCDs provide stable and competitive performance relative to several established clustering methods within the evaluated regimes, while remaining largely parameter-free. We also discuss computational trade-offs and identify the practical regimes in which the method is most effective.
\end{abstract}


\section{Introduction and Motivation}

Clustering is a fundamental statistical method 
for partitioning observations into groups based on similarity or proximity. 
It is an unsupervised learning method
that plays a crucial role in machine learning and pattern recognition.
The primary goal is to group a dataset so that points in the same group (called a cluster) are similar. 
Clustering faces challenges like determining the optimal number of clusters and handling moderate to high-dimensional data. 
Driver and Kroeber first introduced cluster analysis in 1932 \cite{driver1932quantitative}, 
applying it to anthropology. 
Early clustering methods focused on $k$-means and hierarchical methods,
which are the fundamentals of cluster analysis \cite{sorensen1948method, macqueen1967classification}. 
Since then, clustering methods have evolved substantially and are widely used across various disciplines.
Common examples include partitioning customers into distinct groups for better market strategy \cite{wedel2000market}, 
grouping genes that have a similar physiological expression in biology \cite{eisen1998cluster}, 
enhancing image processing by dividing and re-grouping \cite{comaniciu2002mean}, 
and predicting human behavior based on their communities and social ties \cite{girvan2002community}. 
Despite significant advancements, 
clustering remains challenging due to issues such as determining the optimal number of clusters \cite{milligan1985examination}, 
scalability to large datasets \cite{zhang1996birch}, 
robustness to outliers and noise \cite{schubert2017dbscan}, 
and handling moderate to high-dimensional data \cite{beyer1999nearest}.

In our previous work, 
we have proposed a graph- and distribution-based clustering method, 
called UN-CCDs \cite{shi2024outlier},
aiming to overcome the limitations of its two predecessors: 
KS-CCDs and RK-CCDs. 
Specifically, 
due to Ripley's $K$ function employed in \textbf{Monte Carlo Spatial Randomness Test} (MC-SRT), 
RK-CCDs suffer sharp performance degradation when the dimensions exceed 5. 
Also, KS-CCDs are not parameter-free, 
an intensity parameter $\delta$ is needed for a KS-type statistic.
We first recap the CCD-based method below.

\textbf{Cluster Catch Digraphs} (CCDs) are a family of graph-based clustering methods,
which were first introduced by DeVinney \cite{devinney2003class} and Marchette \cite{marchette2005random}, 
adapted from a similar classification digraph called \textbf{Class Cover Catch Digraphs} (CCCDs). 
Later, Manukyan and Ceyhan \cite{manukyan2019parameter} advanced this approach further, 
developing two variants using a Kolmogorov-Smirnov (KS)-type statistic and Ripley's $K$ function,
calling them KS-CCDs and RK-CCDs, respectively.
RK-CCDs and KS-CCDs work similarly on clustering.
While both are effective,
RK-CCDs are particularly appealing for being ``almost" parameter-free, 
unlike KS-CCDs, which require an input intensity parameter for the KS-type statistic.

However, despite their advantages,
they break down when the dimensions exceed 5 \cite{shi2024outlier}.
This limitation motivates the need for a new CCD-based clustering method that is usable in higher dimensions,
while remaining parameter-free.
To address this breakdown, 
we developed UN-CCDs to extend the framework's usability into moderate dimensions (e.g., $d \leq 20$).



The main contributions of this work can be summarized as follows. First, we introduce UN-CCDs, a CCD-based clustering method that replaces the Ripley's $K$-based spatial randomness test with a nearest-neighbor-distance (NND) based Monte Carlo procedure, improving robustness in moderate-dimensional settings. Second, we provide a systematic empirical evaluation demonstrating that the proposed method performs comparably to existing CCD variants in low dimensions and offers improved stability as dimensionality increases within the moderate range (e.g., $d \leq 20$). Third, we show that the method is particularly well-suited for clustering problems involving irregular cluster geometry and uniformly distributed background noise, while requiring minimal parameter tuning. Finally, we clarify the practical scope of CCD-based clustering by identifying the regimes---moderate sample sizes and dimensions---where such methods are most effective.

Despite its advantages, the proposed method has several limitations that define its appropriate scope of application. First, the $O(n^3)$ computational complexity restricts its use to datasets of moderate size, making it less suitable for large-scale or streaming data settings. Second, while UN-CCDs improve robustness relative to existing CCD variants, their effectiveness still degrades in high-dimensional regimes (e.g., $d \gg 20$), where distance-based measures such as nearest-neighbor distances become less informative. Third, the method is primarily designed to handle uniformly distributed background noise rather than general forms of structured outliers, and its performance may vary in datasets with heterogeneous noise patterns or highly overlapping clusters. Finally, although the method is largely parameter-free, it requires the selection of a significance level for the spatial randomness test, which may need adjustment depending on data characteristics. These limitations suggest that UN-CCDs are best viewed as a method tailored to moderate-dimensional, moderate-scale clustering problems with geometric structure and background noise.

The article is organized as follows: In Section \ref{sec:prelim_Clustering}, 
we review clustering methods from several major categories and summarize the mathematical foundations of KS-CCDs and RK-CCDs. 
Section \ref{sec:UN-CCDs} discusses the limitations of RK-CCDs and then presents the motivation and algorithmic formulation of UN-CCDs. 
Sections \ref{sec:Simulation_Clustering} and \ref{sec:Real-Data_clustering} report simulation and benchmark-data results, highlighting the behavior of UN-CCDs relative to RK-CCDs, KS-CCDs, and standard baseline methods within the evaluated settings.

To help readers navigating the specialized terminology used throughout this paper, 
we enumerate a list of acronyms and their full terms below.

\begin{table}[htb]
  \centering
  \resizebox{\textwidth}{!}{\begin{tabular}{|c|c|}
  \hline
  \textbf{Abbreviation} & \textbf{Full Term} \\ \hline
  CCDs & Cluster Catch Digraphs\\ \hline
  RK-CCDs & The CCDs based on the Ripley's $K$ function\\ \hline
  KS-CCDs & The CCDs based on a KS-type statistic\\ \hline
  UN-CCDs & Uniformity- and Neighbor-based CCDs\\ \hline
  MC-SRT & Monte Carlo Spatial Randomness Test\\ \hline
  HPP & Homogeneous Poisson Process\\ \hline
  CSR & Complete Spatial Randomness\\ \hline
  NND & Nearest Neighbor Distance\\ \hline
  ARI & Adjusted Rand Index \\ \hline
  Sil & Silhouette Index \\ \hline
  SR  & The Success Rate to detect the number of clusters \\ \hline
  \end{tabular}}
\end{table}

\section{Preliminaries}
\label{sec:prelim_Clustering}

\subsection{Cluster Methods in Literature}

In literature, cluster-based methods have been classified into several subgroups, 
known as partitional, hierarchical, density-based, grid-based, fuzzy-based, model-based, and distribution-based clustering methods.
We provide a short summary of these methods below. \cite{wang2019survey, yin2024rapid, ezugwu2022comprehensive}.

\textbf{Partitioning clustering} methods iteratively partition data into a predefined number ($k$) of clusters, often minimizing within-cluster distances \cite{wang2019survey, yin2024rapid}. 
They are generally efficient but require $k$ as input and are sensitive to outliers \cite{yin2024rapid}. Classic examples include \textbf{$K$-Means} \cite{gareth2013introduction}, \textbf{$K$-Means++} \cite{arthur2006k}, \textbf{MacQueen} \cite{macqueen1967classification}. 
\textbf{PAM} \cite{kaufman2009finding} is more robust to outliers, but computationally expensive \cite{yin2024rapid}. More recent methods, 
such as \textbf{CLARA} \cite{kaufman2009finding} and \textbf{CLARANS} \cite{ng2002clarans}, address the computational cost of PAM on large datasets.

\textbf{Hierarchical clustering} methods construct a dendrogram with agglomerative (bottom-up) or divisive (top-down) methods, 
without the need to pre-define the number of clusters. 
Those methods offer interpretability but can be computationally costly for large datasets \cite{ran2023comprehensive}. Classic examples include the \textbf{Ward's minimum variance method} \cite{ward1963hierarchical} and \textbf{minimal spanning tree} \cite{MST}, 
but they are computationally intensive and struggle with arbitrarily-shaped clusters \cite{zhang2013advancements}. 
More recent methods, such as \textbf{CURE} \cite{cure}, \textbf{CHAMELEON} \cite{karypis1999chameleon}, and \textbf{ROCK} \cite{guha2000rock} aim to overcome these limitations.

\textbf{Density-based clustering} methods identify clusters as high-density regions, 
separating them by low-density areas. 
They typically do not require $k$ as an input parameter, 
capable of finding arbitrarily shaped clusters,
and are robust to outliers \cite{ester1996density}. 
A typical classic example is \textbf{DBSCAN} \cite{ester1996density}, but its performance is sensitive to the values of two input parameters (\textbf{Eps} and \textbf{MinPts}), 
and has limitations on clusters with varying intensities. Later methods were developed to solve one or two disadvantages of DBSCAN, including \textbf{OPTICS} \cite{ankerst1999optics}, \textbf{DBCLASD} \cite{xu1998distribution}, and \textbf{DENCLUE} \cite{hinneburg1998efficient}.

\textbf{Graph-based clustering} methods partition nodes (points) by leveraging a graph's connectivity structure, aiming to identify sub-graphs for latent clusters. 
A key advantage of those methods is their ability to identify non-convex clusters where vector-based approaches may fail \cite{saxena2017review}. 
Well-known classic examples include the \textbf{Minimal Spanning Tree} (MST) \cite{MST}, 
\textbf{spectral clustering} \cite{ng2001spectral}, 
but these methods are either sensitive to outliers or computationally intensive. 
More recent and scalable methods include \textbf{Louvain} \cite{blondel2008fast} and \textbf{Leiden} \cite{traag2019louvain}, 
which are designed for large datasets. 
We also have deep learning methods like \textbf{Structural Deep Clustering Network (SDCN)} \cite{bo2020structural}. 
This paper focuses on CCD-based methods, 
specifically RK-CCDs and KS-CCDs \cite{manukyan2019parameter}, 
which will be discussed in detail shortly. In Section \ref{sec:Real-Data_clustering}, 
we will use real-life datasets to compare UN-CCDs with some commonly used graph-based methods, 
including MST, spectral clustering, and Louvain.
We will discuss the mechanisms and setup of those methods in Section \ref{sec:Real-Data_clustering}.

Recent work in graph clustering has focused on tackling the challenges of moderate to high-dimensional data and developing more robust methods. 
The ``curse of dimensionality" poses a significant challenge for traditional clustering algorithms,
as distances between any pairs of points become meaningless as dimension increases.
UN-CCDs address this by employing a spatial randomness test with several key enhancements compared to the one employed in RK-CCDs,
making it more robust to higher dimensional data.
Recent research in this area also focuses on moderate to high-dimensional data and using robust statistics to identify and handle outliers that can negatively impact performance. 
Methods like \textbf{RSC-SGM} (Robust Spectral Clustering for Sub-Gaussian Mixtures) construct a Gaussian-kernel similarity graph, 
apply a denoising step to the affinity matrix, 
and use a rounding procedure to achieve accurate and automatic clustering and outlier detection \cite{srivastava2023robust}.
\textbf{NA-COV} (Network-Adjusted Covariates) integrates high-dimensional node covariates with network structure to improve community detection in sparse graphs \cite{hu2024network}.
\textbf{BayesCov-SBM} (Bayesian Community Detection with Covariates) uses a joint model called Bayesian stochastic block model that incorporates both network structure and high-dimensional node covariates for community detection \cite{shen2025bayesian}.

Graph-based clustering methods, 
including the CCD-based framework, 
leverage the digraph's connectivity and relative distance for clustering. 
A key advantage of CCD-based methods is identifying non-convex clusters \cite{manukyan2019parameter}, 
where vector-based approaches like $K$-Means and Hierarchical clustering may fail. 
Additionally, unlike density-based methods like DBSCAN that identify the ``seeds" of clusters in high-density regions exceeding a density threshold, 
most CCDs do not require the user to define such a threshold. 
Instead, CCDs create a digraph where arcs are determined by a spatial randomness test for each data point, 
effectively identifying local clusters. 
CCD-based methods also differ from modularity-based graph methods, such as Louvain, 
which partitions a network by maximizing a measure of community structure. 
While Louvain and other similar methods are designed for large networks, 
CCDs are particularly focused on using geometric properties of the data itself to define the graph structure and clusters.  

Several modern categories of clustering, including grid-based, model-based, and fuzzy-based clustering methods were developed to address specific challenges like large-scale data, statistical rigor, and ambiguous cluster boundaries.

\textbf{Grid-based clustering} methods partition the data space into a grid structure and perform clustering on the grid cells. 
Their time/space complexity depends on grid size rather than data size, 
making them fast for large datasets \cite{wang1997sting}. 
The clustering quality relies on grid granularity, 
and they may struggle with irregular cluster boundaries \cite{yin2024rapid}. 
Examples include \textbf{STING} \cite{wang1997sting}, \textbf{WaveCluster} \cite{sheikholeslami2000wavecluster}, \textbf{DClust} \cite{zhang2005clustering}, and \textbf{CLIQUE} \cite{agrawal1998automatic}.

\textbf{Model-based clustering} methods assume data originates from a mixture of probability distributions, these methods fit models to identify clusters \cite{ran2023comprehensive}. 
They offer flexibility in cluster shape/size and support ``soft" assignments but can be computationally intensive and are sensitive to model/parameter choices \cite{yin2024rapid}. 
Examples include \textbf{GMMs} \cite{rasmussen1999infinite}, \textbf{LDA} \cite{blei2003latent}, \textbf{HMM} \cite{rabiner1986introduction}, \textbf{COBWEB} \cite{fisher1987knowledge}, \textbf{pdfCluster} \cite{azzalini2013clustering}, and \textbf{SOM} \cite{vesanto2000clustering}.

\textbf{Fuzzy Clustering} is known as ``soft" clustering, 
these methods allow data points to belong to multiple clusters with varying degrees of membership \cite{ezugwu2022comprehensive}, 
useful for ambiguous or overlapping boundaries \cite{kaufman2009finding}. 
Optimization can be computationally intensive \cite{ezugwu2022comprehensive}. 
Examples include \textbf{Fuzzy C-means} (FCM) \cite{dunn1973fuzzy}, which is sensitive to initialization and outliers \cite{saxena2017review}; the more robust \textbf{Possibilistic-fuzzy C-means} (PFCM) \cite{pal2005possibilistic, krishnapuram1996possibilistic}; and \textbf{Kernelized Fuzzy C-means} (KFCM) for non-linear boundaries \cite{zhang2004novel}.

\subsection{Cluster Catch Digraphs}
\label{sec:CCDs}
Our clustering methods are based on Cluster Catch Digraphs (CCDs), 
where vertices represent the points of a given dataset $\mathrm{X}$,
and arcs are determined by the spherical ball (covering ball) centered at each point. 
Class Cover Catch Digraphs (CCCDs) \cite{marchette2003classification} are the prototypes of CCDs, 
originally designed for supervised classification.
CCCDs aim to distinguish a target class from a non-target class by covering the target class points with a minimum number of balls while excluding the non-target points \cite{marchette2005random}. 
A CCCD is constructed by creating covering balls around each point, 
forming arcs between points when one lies within the covering ball of another. 
Two variants, pure-CCCDs (P-CCCDs) and random walk-CCCDs (RW-CCCDs), differ in how the radius of each covering ball is determined \cite{manukyan2016classification}.
CCDs are an unsupervised adaptation of CCCDs \cite{marchette2005random},
initializing by constructing a covering ball for each point. 
To reduce the covering complexity in CCDs and avoid overfitting, 
we retain only a subset of covering balls that optimally covers all (or most) points in a dataset.
One way is finding a \textbf{minimum dominating set} (MDS) of the CCD.
Although finding the MDS of a digraph is an NP-Hard problem \cite{karp1972reducibility},
one can find an approximation of MDS efficiently with the Greedy Algorithm,
which operates in $O(|V_0|^2)$ time \cite{chvatal1979greedy, hochbaum1982approximation}. 
The Greedy Algorithm finds an approximate MDS 
by iteratively selecting the vertex with the highest outdegree 
at each iteration and storing this point in a set,
and removing its closed neighborhood (in the graph) from the uncovered vertices until all vertices are covered.
When the algorithm terminates,
the stored vertices are retained for the (approximate) MDS.
Manukyan and Ceyhan proposed two variants of the Greedy Algorithm for CCDs \cite{manukyan2019parameter}: 
the first variant prioritizes vertices closer to cluster centers, 
while the second maximizes a score function at each step.
We provide these methods in Algorithms \ref{alg:greedy-outdegree-orig_digraph} and \ref{alg:greedy-score-func},
respectively for completeness,
because they are non-standard in their greediness and also for future reference.

\begin{algorithm}[htb]
\setcounter{algocf}{0}
\SetAlgorithmName{Greedy Algorithm}{}{}
\caption{A greedy algorithm to find an approximate MDS of CCDs (which \textbf{does not} update the digraph after each iteration). 
$V$ and $A$ are the initial vertex and edge sets of the digraph $D$, respectively.  
$\bar{N}(v)$ is the closed neighborhood of a vertex $v$ (i.e., the points covered by the covering ball centered at $v$, including $v$ itself).
At each iteration,
$V_{temp}$ represents the uncovered vertices,
and $v_{temp}$ is the vertex with largest outdegree.}
\label{alg:greedy-outdegree-orig_digraph}
\KwData{A digraph $D=(V(D)=\mathrm{X},A(D))$ where $\mathrm{X} = \{x_1,x_2,...,x_n\}$}\\
\KwResult{An approximate minimum dominating set $\hat{S}$}\\
\vspace{8pt}
\nl\textbf{Initialization:} $V_{temp} \gets V(D)$, $\hat{S} \gets \emptyset$\\
\vspace{5pt}
\nl\While{$V_{temp} \neq \emptyset$}{
\nl $v_{temp}\gets \arg\max_{v \in V \symbol{92} \hat{S}}\{d^*_{out}(v)\}$\Comment*[r]{$d^*_{out}(v)$:the outdegree of $v$ in $D$}
\nl $V_{temp} \gets V_{temp} \symbol{92} \bar{N}(v_{temp})$\;
\nl $\hat{S} \gets \hat{S} \cup \{v_{temp}\}$\;
}
\end{algorithm}

\begin{algorithm}[htb]
\SetAlgorithmName{Greedy Algorithm}{}{}
\caption{A greedy algorithm to find an approximate MDS which is greedy
in a score function $sc(v)$ at each iteration (digraph is updated after each iteration).
$D_{sub}(S)$ is the induced sub-digraph of vertex set $S$ from a digraph $D$.
}
\label{alg:greedy-score-func}
\KwData{A digraph $D=(V(D),A(D))$ for a given dataset $\mathrm{X} = \{x_1,x_2,...,x_n\}$}\\
\KwResult{An approximate minimum dominating set $\hat{S}$}\\
\vspace{8pt}
\nl\textbf{Initialization:} $V_{temp} \gets V(D)$, $\hat{S} \gets \emptyset$\\
\vspace{5pt}
\nl\While{$V_{temp} \neq \emptyset$}{
\nl $v_{temp}\gets \arg\max_{v \in V(D)}\{sc(v)\}$\Comment*[r]{$sc(v)$:the score function for $v$}
\nl $V_{temp} \gets V_{temp} \symbol{92} \bar{N}(v_{temp})$\;
\nl $\hat{S} \gets \hat{S} \cup \{v_{temp}\}$\;
\nl $D \gets D_{sub}(V_{temp})$\;
}
\end{algorithm}

\subsubsection{Cluster Catch Digraphs Based on a KS-Type Statistic}
\label{sec:KS-CCD}
CCDs, introduced by Devinney \cite{devinney2003class}, 
are an adaptation of CCCDs \cite{marchette2003classification} for unsupervised clustering. 
Given an unlabeled dataset $\mathrm{X} = \{x_1, x_2, ..., x_n\} \subset \mathbb{R}^d$, 
where points are drawn from a mixture distribution, 
the objective is to determine the number of clusters and the optimal data partitioning. 
Unlike CCCDs, 
CCDs determine the radius of each covering ball $B(x_i, r_{x_i})$ (centered at $x_i$ with radius $r_{x_i}$) using a Kolmogorov-Smirnov (KS)-type statistic, 
which measures the deviation of the local intensity around each point $x_i$ from a null model such as the \textbf{Complete Spatial Randomness} (CSR, where points are distributed uniformly within a given area) \cite{marchette2005random}, the KS-type statistic for a point $x_i$ is defined as:
\begin{equation}
  T_{KS}(x_i, r) = F_{RW}(x_i, r) - F_0(x_i, r),
\end{equation}
where $F_{RW}(x_i, r)$ represents the number of points within the covering ball $B(x_i, r)$, 
and $F_0(x_i, r)$ is the expected number of points under the null model, 
typically given by $F_0(x_i, r) = \delta r^d$ under the assumption of CSR, 
where $\delta$ is a density parameter and $d$ is the dimension of the space.

The radius $r_{x_i}$ of a covering ball $B(x_i, r_{x_i})$ is chosen to maximize its KS-type statistic:
\begin{equation}
r_{x_i} = \arg\max_{r \geq 0} \{ T_{KS}(x_i, r) \},
\end{equation}
which ensures that the points covered by $B(x_i, r_{x_i})$ (including $x_i$) form a significant local cluster.

Once the optimal radii are determined for all covering balls, 
a digraph for $\mathrm{X}$ is constructed,
denoted as $D = (V(D), A(D))$, 
where $V(D) = \mathrm{X}$, 
and there is an arc from $x_i$ to $x_j$ ($x_ix_j \in A(D)$) if $x_j \in B(x_i, r_{x_i})$.
Once the digraph $D$ is constructed,
the weakly connected components of $D$ can be taken as clusters.

In order to prevent the impact of noise clusters,
one may want to reduce the ``complexity" of the covering balls.
To achieve this goal,
we apply the Greedy Algorithm \ref{alg:greedy-outdegree-orig_digraph} to find an approximate MDS, 
denoted as $\hat{S}$, 
which selects a subset of covering balls,
representing the clusters more efficiently \cite{marchette2005random}. 
The algorithm iteratively selects the vertex with the largest outdegree (i.e., the vertex with covering ball covering the most points) until all points are covered.

To further reduce the complexity of the cluster cover, 
an \textbf{intersection graph} $G_{MD} = (V_{MD}, E_{MD})$ is constructed, 
where $V_{MD} = \hat{S}$, 
and an edge exists between $u, v \in \hat{S}$ if their covering balls cover some common points, i.e., $\bar{N}(u) \cap \bar{N}(v) \neq \emptyset$. 
The Greedy Algorithm \ref{alg:greedy-score-func} is then applied to $G_{MD}$ to prune the MDS $\hat{S}$ further, 
yielding a reduced set $\hat{S}(G_{MD})$ that represents the center of latent clusters,
and an initial partition $P = \{P_1, P_2, ..., P_k\}$ for $\mathrm{X}$ can be constructed based on $\hat{S}(G_{MD})$.

Finally, to remove noise or outlying clusters,
Manukyan and Ceyhan \cite{manukyan2019parameter} designed an automatic refinement process based on \textbf{silhouette index} (a metric measures the quality of clustering) \cite{gan2020data}.
This process first ranks the partitions in $P$ in a decreasing order based on their size. 
Starting from the first two, 
it adds partitions incrementally as valid clusters until the average silhouette index of the entire dataset is maximized,  
indicating the optimal number of clusters is reached.
In summary, the refinement process works on a set of candidate partitions (or clusters) and the final, 
selected ones are the finalized clusters, 
whose covering balls are the \textbf{dominating covering balls}.

During the above automatic refinement process,
re-partitioning occurs after each incremental addition of a new partition, 
in order to re-evaluate the overall average silhouette index.
In the re-partitioning step,
points are assigned to the existing partition with the smallest relative dissimilarity measure $\rho(x_i, B(x_j, r_{x_j}))$, which is defined as follows,
\begin{equation}
\rho(x_i, B(x_j, r_{x_j})) = \frac{d(x_i, x_j)}{r_{x_j}},
\end{equation}
where $d(x_i, x_j)$ is any distance or dissimilarity measure and $B(x_j, r_{x_j})$ is the covering ball for an existing partition during this process.

\subsubsection{Cluster Catch Digraphs Based on Ripley's \textit{K} Function}
\label{sec:RK-CCD}
Although the KS-CCD method is robust to noise clusters thanks to the employment of the silhouette index, 
it still has some limitations.
For example, the KS-CCD method is not able to capture the spatial distributions of each cluster,
and it needs an input density parameter $\delta$ to determine the radius of each covering ball \cite{manukyan2019parameter}. 
To address those limitations, 
instead of using the KS-type statistic, 
Manukyan and Ceyhan \cite{manukyan2019parameter} applied the Ripley's \textit{K} function \cite{ripley1976second}, denoted as $K(t)$, 
to construct a test determining whether the points within each covering ball follow a \textbf{homogeneous Poisson process} (HPP, i.e., Complete Spatial Randomness). 
This test is referred to as the \textbf{Monte Carlo Spatial Randomness Test} (MC-SRT) with Ripley's \textit{K} function.

More specifically, given a dataset $\mathrm{X}$ consisting of \textit{i.i.d} points $\{x_1,x_2,...,x_n\}$ in a region $R \subset \mathbb{R}^d$,
the Ripley's $K$ function measures the spatial distribution of $\mathrm{X}$, 
revealing whether the points of $\mathrm{X}$ are clustered, dispersed, or randomly distributed.
It is defined as follows \cite{dixon2012ripley},
\begin{equation}\label{equ:K-function_tech3}
  K(t) = \lambda^{-1} E[\text{the number of additional pairs of points within distance $t$}],
\end{equation}
where $\lambda$ is the density of the point set.

$K$ function can be applied to describe certain point patterns of $\mathrm{X}$ and the most common application is to test CSR \cite{dixon2012ripley}, 
where the test statistic is the estimator of $K(t)$, 
denoted as $\widehat{K}(t)$, which is proportional to the ratio of the actual number of pairs whose distance is less than $t$ against the intensity of the point set.
For more details,
please refer to Manukyan and Ceyhan's work \cite{manukyan2019parameter}.

The RK-CCD method sets the radius of each covering ball to the largest value for which the points inside the ball fail to reject the null hypothesis of CSR,
ensuring that the points encapsulated conform to a CSR pattern. 
The primary difference between RK-CCDs and KS-CCDs is in the determination of the radii, 
which is now based on Ripley's \textit{K} function rather than the KS-type statistic. 
Subsequent steps, such as reducing cover complexity using an approximate MDS, 
and refining the clustering result by maximizing the silhouette index, 
remain the same as in KS-CCDs.

A variant of RK-CCDs introduced by Manukyan and Ceyhan allows for the detection of clusters with arbitrary shapes by considering the connected components of the intersection graph $G_{MD}$ as the final clusters instead of using the MDS \cite{manukyan2019parameter}.

\section{Cluster Catch Digraphs using the Nearest Neighbor Distance}
\label{sec:UN-CCDs}
\subsection{The Limitations of RK-CCDs for Clustering}
\label{sec:RK-CCDs_limits}
In our previous work,
we proposed the U-MCCD and SU-MCCD methods for outlier detection \cite{shi2024outlier}, 
where SU-MCCDs are the ``flexible" versions of U-MCCD,
adaptable for arbitrarily shaped clusters.
As their first step, 
both methods employ RK-CCDs for clustering.
However, the Monte Carlo analysis reveals that outlier detection with U-MCCDs and SU-MCCDs exhibit high \textbf{False Positive Rates} (FPRs) 
for regular points (falsely identifying regular points as outliers) when dimension increases,
especially when dimensions exceed 10.
We have shown that this degradation is attributed to the limitations of RK-CCDs in higher dimensional spaces, 
whose dominating covering balls are not sufficiently large for latent clusters \cite{shi2024outlier}.

Specifically, RK-CCDs present two key limitations: 
(1) The initial points in the MC-SRT are not random, 
as they are always the centers of the covering balls, 
leading to inaccurate testing results, 
especially in moderate to high-dimensional space;
(2) In MC-SRT, the point-wise confidence band for $K(t)$ is constructed on fixed $t$ values \cite{manukyan2019parameter}, 
which is not appropriate, 
especially as the dimensionality of the dataset increases 
and the distances between any two points become arbitrarily large.

\subsection{The MC-SRT with the Nearest Neighbor Distance}
\label{sec:MC-SRT_NND}
To avoid the limitations of RK-CCDs, 
we have introduced a new CCD approach for several new outlier detection methods and scores \cite{shi2024outlier, shi2025outlierscores}.
This new CCD conducts MC-SRT with the \textbf{Nearest Neighbor Distance} (NND) instead of the Ripley's \textit{K} function.

Given a dataset $\mathrm{X} = \{x_1,x_2,...,x_n\}$,
the mean NND of $\mathrm{X}$ is $\bar{\mathrm{d}}=\frac{\sum_{i=0}^{n}\mathrm{d}_i}{n}$, 
where $\mathrm{d}_i$ represents the distance of $x_i$ to its nearest neighbor.
We aim to compare $\bar{\mathrm{d}}$ with the expected mean NND (denoted as $\mu_{\mathrm{d}}$) under the assumption of CSR.
A normality test offers a convenient way to obtain $p$-values for $\bar{\mathrm{d}}$, 
as both $\mu_{\mathrm{d}}$ and $\sigma_{\bar{\mathrm{d}}}$ 
(the standard deviation of $\bar{\mathrm{d}}$ under the assumption of CSR) can be derived analytically \cite{clark1954distance, clark1979generalization}.
However, when the sample size $n$ is small, 
the distribution of $\bar{\mathrm{d}}$ is skewed, 
making the normality test unfeasible \cite{besag1977simple}.
Fortunately, a Monte Carlo test proposed by Besag and Diggle \cite{besag1977simple} provides an alternative,
where the $p$-values are determined by the empirical quantiles.

Consequently, we have proposed another CCD approach based on the MC-SRT with NND,
presented as Algorithm \ref{alg:NNtest}.
Specifically, we made several key enhancements to boost the robustness and reliability of the new CCDs: 
(1) We conduct MC-SRT with both median and mean NND, 
applying a Holm's Step-Down Procedure for correction \cite{witten2013introduction}, 
which improves the accuracy of the test substantially;
(2) Our previous work shows that larger covering balls for each point are preferred for higher dimensional space \cite{shi2024outlier}. 
Therefore, we offer the option to test the candidate radius values in a descending order,
stopping when the current radius value is not rejected in MC-SRT.
(3) The MC-SRT is lower-tailed as the upper tail indicates the point pattern is significantly ``regular", 
which is less relevant to our interest.
(4) The center points are excluded from the test as they are not random.

\begin{algorithm}[htb]
\KwData{A hypersphere in $\mathbb{R}^d$ with radius $r$ covering i.i.d point set $\mathcal{X}_{sub}$ of size $n_{sub}$ from $\mathcal{X}$}\;

\KwResult{Decision on CSR rejection for $\mathcal{X}_{sub}$ at level $\alpha$}\;

\textbf{Algorithm Steps:}
Compute mean $\bar{\mathcal{D}}$ and median $\widetilde{\mathcal{D}}$ of NND in $\mathcal{X}_{sub}$, scaled by $r$\;

Simulate $m$ sets within a unit sphere in $\mathbb{R}^d$, each of size $n_{sub}$, under CSR\;

Calculate mean $\{\bar{\mathcal{D}}_i\}$ and median $\{\widetilde{\mathcal{D}}_i\}$ NNDs for simulations\;

Determine empirical p-values $p_1$ for $\bar{\mathcal{D}}$ and $p_2$ for $\widetilde{\mathcal{D}}$, then order $p_{(1)} \leq p_{(2)}$\;

Reject CSR for $\mathcal{X}_{sub}$ if $p_{(1)} \leq \alpha/2$ or $p_{(2)} \leq \alpha$ using Holm's Step-Down Procedure \cite{witten2013introduction}\;

\caption{Spatial Randomness Monte Carlo Test (MC-SRT) Using NND}
\label{alg:NNtest}
\end{algorithm}

This localized, NND-based approach natively handles background noise contamination. 
We explicitly distinguish this from general outlier detection in the simulation study of Section \ref{sec:simulation_W_noise},
as our framework specifically targets uniformly distributed background noise rather than extreme point-source anomalies. 
Decoupling noise removal into a preprocessing phase typically relies on global distance or density thresholds, 
risking the permanent deletion of valid points. 
UN-CCDs resolve this by filtering background noise through their geometric mechanics: random background points fail to form significant covering balls, yield low outdegrees, 
and are naturally excluded during MDS construction. 
This dynamically isolates spatial noise based strictly on local topology, 
eliminating parameter dependence.

The new clustering approach based on CCDs and NND is called \textbf{Uniformity- and Neighbor-based CCD} (UN-CCD) clustering method, 
which is presented in Algorithm \ref{alg:UN-CCDs_2}. 
It is worth noting that the only difference among KS-CCDs, RK-CCDs, and UN-CCDs is how to determine the radius of covering balls.
The time and space complexities of UN-CCDs for clustering are $O((N+d)n^2 + n^3)$ and $O(n^2)$ respectively, which we have proven in our previous work.

We summarize the time and space complexity of RK-CCDs, KS-CCDs, and UN-CCDs in Table \ref{tab:space_time_CCD_clustering}.
When $N$ and $d$ are fixed, KS-CCDs and UN-CCDs have the time complexity of $O(n^3)$,
while RK-CCDs run in $O(n^3 \log n)$,
slower than the other two.
All the three methods have the same space complexity of $O(n^2)$.

\begin{algorithm}[htb]
\KwData{$\alpha$, dataset $\mathrm{X} = \{x_1,x_2,...,x_n\}$}\;
\KwResult{Cluster centers of $\mathrm{X}$}\;
\vspace{8pt}
\textbf{Algorithm Steps:}\\
\nl \ForEach{$x_i \in \mathrm{X}$}{
\nl Calculate distances $D(x_i) = \{d(x_i, x_j) | x_j \in \mathrm{X}, x_i \neq x_j\}$\;
\vspace{5pt}
\nl \ForEach{distance $r_{(j)}$ in $D(x_i)$ sorted ascending}{
\nl Perform the MC-SRT with NND (Algorithm \ref{alg:NNtest}) on the $\mathrm{X} \cap B(x_i, r_{(j)})$\;
\vspace{5pt}
\nl \If{test rejected at level $\alpha$}{
\nl Set $r_{x_i} = r_{(j-1)}$; \textbf{break}\;
}\vspace{5pt}
}\vspace{5pt}
}\vspace{5pt}
\nl Construct a CCD, $D=(V,A)$ using the pre-determined radii\;
\nl Find the approximate minimum dominating set $\hat{S}(V)$ in $D$ with the Greedy Algorithm \ref{alg:greedy-outdegree-orig_digraph}\;
\nl Create intersection graph $G_{MD}=(V_{MD},E_{MD})$ with $\hat{S}(V)$\;
\nl Find an approximate minimum dominating set $\hat{S}(G_{MD})$ in $G_{MD}$ using the Greedy Algorithm \ref{alg:greedy-score-func} with a score function measuring the number of points covered, stops when the average silhouette index $Sil(P)$ is maximized\; 
\nl Output $\hat{S}(G_{MD})$ as cluster centers\;
\caption{(\textbf{UN-CCD Clustering method}) Cluster Catch Digraphs based on the Nearest Neighbor Distance (NND). $\alpha$ is the level of the Monte Carlo test with NND.}
\label{alg:UN-CCDs_2}
\end{algorithm}

\begin{table}[htb]
  \centering
  \begin{tabular}{|c|c|c|}\hline
  \textbf{Algorithms} & \textbf{Time Complexity} & \textbf{Space Complexity} \\ \hline
  KS-CCDs & $O(n^3+n^2(d+\log n))$ & $O(n^2)$ \\ \hline
  RK-CCDs & $O(n^3(\log n+ N)+n^2d)$ & $O(n^2)$ \\ \hline
  UN-CCDs & $O((N+d)n^2+n^3)$ & $O(n^2)$ \\ \hline
\end{tabular}
\caption{The time/space complexity of all CCD-based methods.}
\label{tab:space_time_CCD_clustering}
\end{table}

\begin{figure}[H]
    \centering
    \begin{subfigure}[t]{0.45\textwidth}
        \centering
        \includegraphics[width=\textwidth]{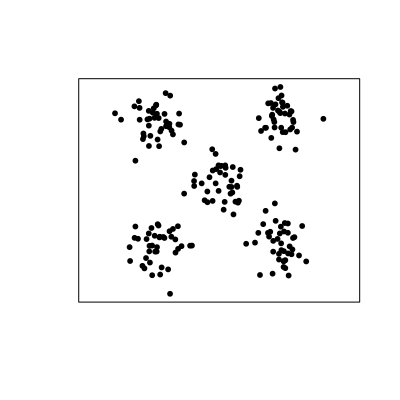}
        \caption{}
        \label{fig:UN-CCD_plot_1}
    \end{subfigure}
    \hfill
    \begin{subfigure}[t]{0.45\textwidth}
        \centering
        \includegraphics[width=\textwidth]{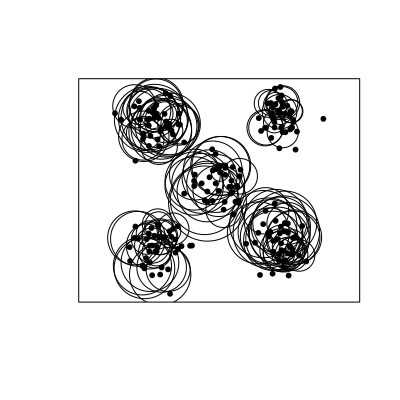}
        \caption{}
        \label{fig:UN-CCD_plot_2}
    \end{subfigure}
    
    \begin{subfigure}[t]{0.45\textwidth}
        \centering
        \includegraphics[width=\textwidth]{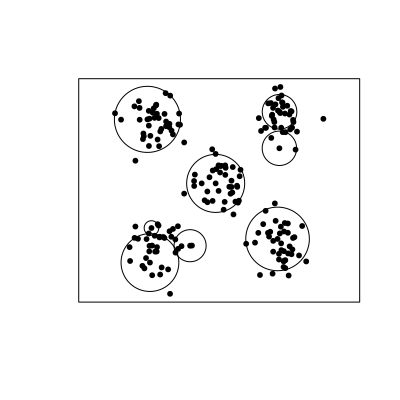}
        \caption{}
        \label{fig:UN-CCD_plot_3}
    \end{subfigure}
    \hfill
    \begin{subfigure}[t]{0.45\textwidth}
        \centering
        \includegraphics[width=\textwidth]{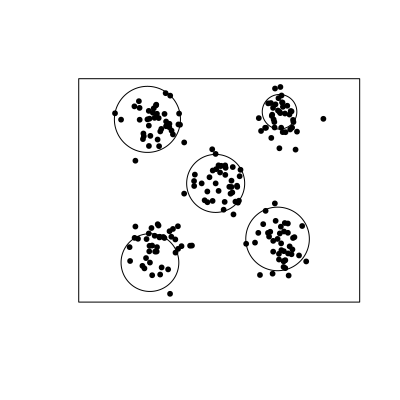}
        \caption{}
        \label{fig:UN-CCD_plot_4}
    \end{subfigure}
    
    \caption{An illustration of clustering with UN-CCDs. \textbf{Top-left:} A dataset consisting of 5 clusters generated from 5 different bivariate normal distributions. \textbf{Top-right:} The covering balls of an approximate MDS obtained by Greedy Algorithm \ref{alg:greedy-outdegree-orig_digraph}. \textbf{Bottom-left:} The covering balls of an approximate MDS of the intersection graph. \textbf{Bottom-right:} The dominating covering balls of the intersection graph that maximize silhouette index $Sil(P)$.}
    \label{fig:UN-CCD_plot}
\end{figure}

The clustering procedure with UN-CCDs for a 2-dimensional dataset consisting of $5$ clusters is illustrated in Figure \ref{fig:UN-CCD_plot}.

In Section \ref{sec:Simulation_Clustering}, we compare the performance of KS-CCDs, RK-CCDs, 
and UN-CCDs with synthetic datasets consisting of multiple clusters. 
We show the advantage of UN-CCDs over RK-CCDs on Gaussian clusters and in settings with moderate sizes and dimensions.

\section{Monte Carlo Experiments for Clustering Based on UN-CCDs}
\label{sec:Simulation_Clustering}

The code, datasets, and simulation results of this section and all the following sections can be found on \url{https://github.com/Rui-Shi/Cluster-Catch-Digraphs-for-Clustering-and-Outlier-Detection/tree/main/CCDs_Outlier_Detection}.

\subsection{Experiment with Uniform Settings}
This section presents a Monte Carlo analysis of CCD-based clustering using synthetic datasets. 
These datasets are generated with varying dimensionality $d$, 
dataset size $n$, 
number of clusters, 
and cluster volumes.

Similar to the design of simulations in our previous work \cite{shi2024outlier},
the analysis begins with datasets consisting of uniform clusters.

To evaluate the method's performance across low to moderate dimensions, 
with $d=10$ and $d=20$ representing the upper limits of our moderate-dimensional testing, 
the varied simulation parameters are enumerated as follows:

\begin{itemize}
\item[\romannumeral1.] Dimensionality ($d$): 2, 3, 5, 10, 20;
\item[\romannumeral2.] Dataset size ($n$): 50, 100, 200, 500;
\item[\romannumeral3.] Number of clusters (\# clusters): 2, 3, 5.
\end{itemize} 

Additionally, all simulated datasets share the following common factors:
\begin{itemize}
\item[\romannumeral1.] Cluster radii is a random variable between 0.8 and 1.2 (uniformly distributed);
\item[\romannumeral2.] Equal cluster sizes (with varying support volumes and intensities due to the changing radii);
\end{itemize}

\begin{figure}[htb]
    \centering
    \begin{subfigure}[t]{0.32\linewidth}
        \centering
        \includegraphics[width=\textwidth]{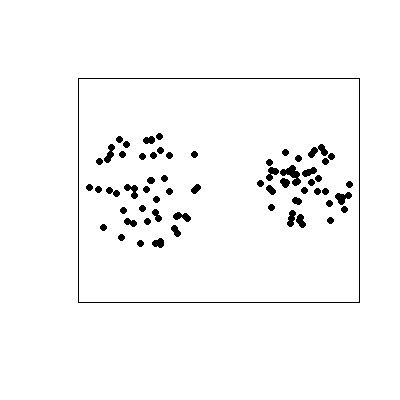}
        \caption{2 clusters}
        \label{fig:2_cls_Apdx}
    \end{subfigure}
    \hfill
    \begin{subfigure}[t]{0.32\linewidth}
        \centering
        \includegraphics[width=\textwidth]{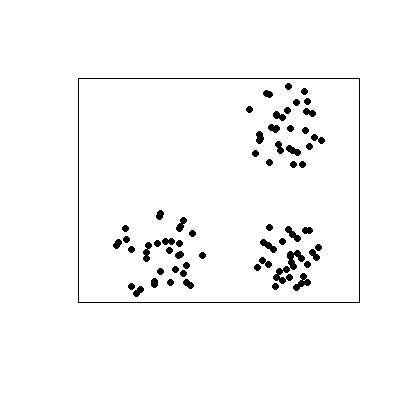}
        \caption{3 clusters}
        \label{fig:3_cls_Apdx}
    \end{subfigure}
    \hfill
    \begin{subfigure}[t]{0.32\linewidth}
        \centering
        \includegraphics[width=\textwidth]{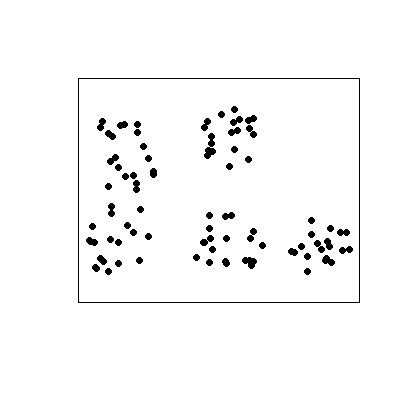}
        \caption{5 clusters}
        \label{fig:5_cls_Apdx}
    \end{subfigure}
    \caption{Realizations of the simulation settings with 2, 3, and 5 uniform clusters in $\mathbb{R}^2$.}
    \label{fig:Demo_2d_Uniform_Cls_Apdx}
\end{figure}

Additionally, the cluster centers are: 
\begin{itemize}
  \item $1^{st}$ cluster: $\bm{\mu_1} = (\underbrace{3,...,3}_{d})$;
  \item $2^{nd}$ cluster: $\bm{\mu_2} = (6,\underbrace{3,...,3}_{d-1})$;
  \item $3^{rd}$ cluster: $\bm{\mu_3} = (6, 6, \underbrace{3,...,3}_{d-2})$;
  \item $4^{th}$ cluster: $\bm{\mu_4} = (3, 5.5, \underbrace{3,...,3}_{d-2})$;
  \item $5^{th}$ cluster: $\bm{\mu_5} = (8.5, \underbrace{3,...,3}_{d-1})$.
\end{itemize}
The minimal distances between the first three cluster centers are 3, 
ensuring well separate clusters.
On the other hand, the distances between the $1^{st}$ and the $4^{th}$,
the $2^{nd}$ and the $5^{th}$ cluster centers are 2.5,
leading in closer clusters,
which may be confused as one cluster by many clustering methods.
This study investigates the performance of the three CCD-based clustering methods for close clusters.
We present three realizations with different number of clusters under 2-dimensional space in Figure \ref{fig:Demo_2d_Uniform_Cls_Apdx}.


Clustering quality is assessed using both internal and external validation measures, 
including the Adjusted Rand Index (\textbf{ARI}) \cite{hubert1985comparing}, 
the average silhouette index (\textbf{Sil}) \cite{rousseeuw1987silhouettes}, 
and the frequency of successfully identifying the correct number of clusters (denoted as success rate or \textbf{SR} for simplicity).
We repeat each simulation 500 times,
and the average measures across these repetitions are recorded.

\begin{figure}[htb]
\centering
    \begin{subfigure}[t]{1\linewidth}
        \centering
        \includegraphics[width=\textwidth]{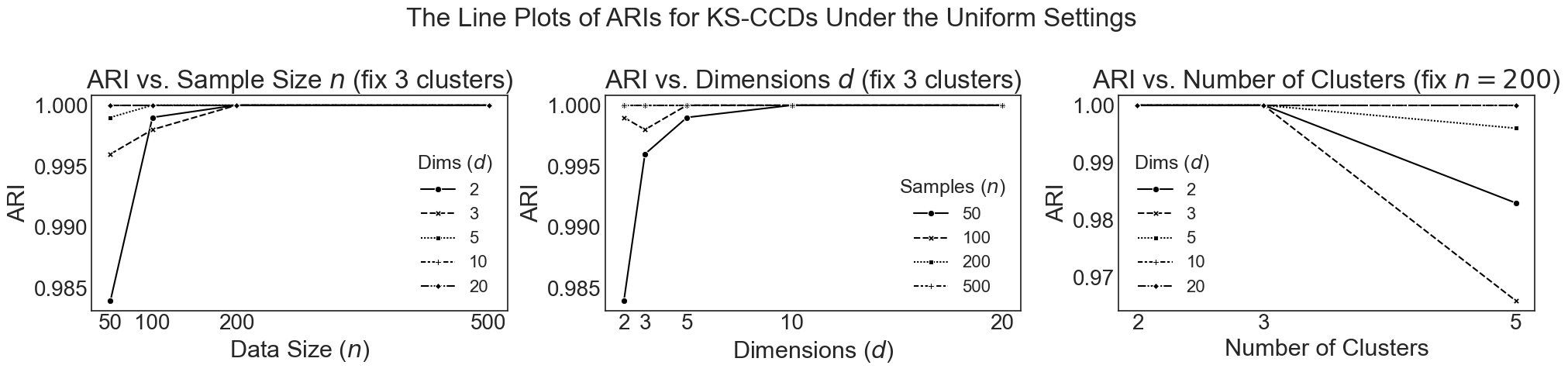}
    \end{subfigure}
\caption{The line plots of the ARIs of KS-CCDs, under the uniform cluster settings.}
\label{fig:Uniform_ARI_KSCCDs_Apdx}
\end{figure}

\begin{figure}[htb]
    \centering
    \begin{subfigure}{1\linewidth}
        \centering
        \includegraphics[width=\linewidth]{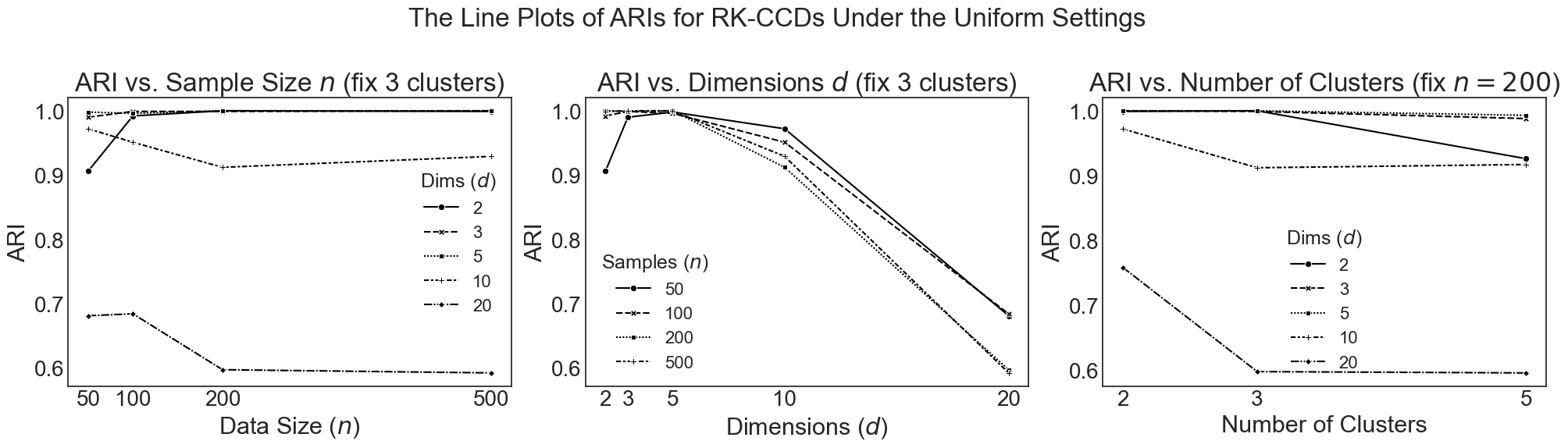}
    \end{subfigure}
    \caption{The line plots of the ARIs of RK-CCDs, under the uniform cluster settings.}
    \label{fig:Uniform_ARI_RKCCDs_Apdx}
\end{figure}

\begin{figure}[htb]
    \centering
    \begin{subfigure}{1\linewidth}
        \centering
        \includegraphics[width=\linewidth]{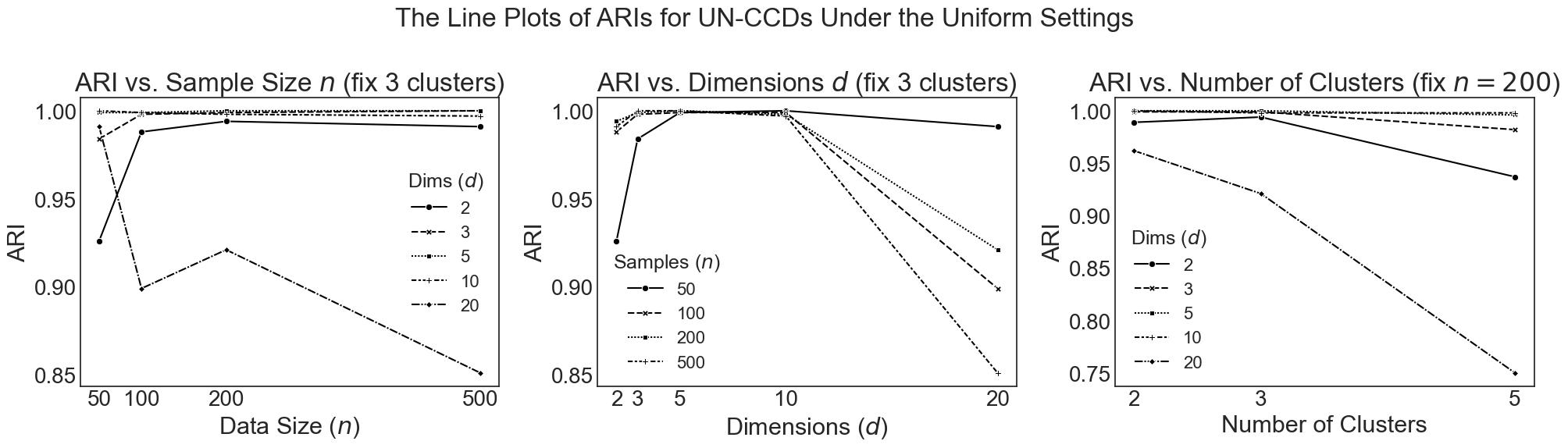}
    \end{subfigure}
    \caption{The line plots of the ARIs of UN-CCDs, under the uniform cluster settings.}
    \label{fig:Uniform_ARI_UNCCDs_Apdx}
\end{figure}

To accommodate the decreased data intensity in higher dimensional datasets,
one may increase the volumes of covering balls as dimensions increase.
A smaller significance level $\alpha$ of MC-SRT (for both RK-CCDs and UN-CCDs) leads to larger radius for the covering balls.
Therefore, we can achieve this objective by tuning the significance level $\alpha$ of MC-SRT (for both RK-CCDs and UN-CCDs),
After some trial-and-error process (not shown here),
we set $\alpha$ to $1\%$ for $d<10$ and $0.1\%$ for $d\geq10$ for RK-CCDs,
which achieves the best overall performance.
For UN-CCDs, 
the $\alpha$ of the MC-SRT (with NND) is set to $\{15\%,10\%,5\%,1\%,0.1\%\}$ as $d$ increases from 2 to 20 for the optimal performance.
Unlike RK-CCDs and UN-CCDs,
KS-CCDs require a density parameter $\delta$ for the KS-type statistic.
To manage the scale of $\delta$ in higher dimensional settings,
we work on the value of $\sqrt[d]{\delta}$ instead,
and select the $\sqrt[d]{\delta}$ maximizing the average silhouette index in each simulation,
presented in Tables \ref{tab:uni_KS-CCDs_den_Apdx} and \ref{tab:Gau_KS-CCDs_den_Apdx}.
The simulation results are summarized in Tables \ref{tab:uni_KS-CCDs_Apdx}, \ref{tab:uni_RK-CCDs_Apdx}, and \ref{tab:uni_UN-CCDs_Apdx}.
For better visualization, we summarize the simulation results in the following line plots (Figures \ref{fig:Uniform_ARI_KSCCDs_Apdx} to \ref{fig:Uniform_ARI_UNCCDs_Apdx}), illustrating the trend of the performance (ARIs) of each CCD-based methods with varying data sizes ($n$), dimensions ($d$), and the number of clusters.
To keep the figures uncluttered,
we fix the number of clusters to 3 for the line plots with varying data sizes and dimensions;
and the data size $n$ is fixed to 200 when we study the performance trend as the number of clusters increases.

\begin{table}[htb]
  \centering
  \footnotesize
  \begin{tabular}{|c|c|c|c|c|c|c|}
  \hline

  \multicolumn{2}{|c|}{Density Parameter: $\sqrt[d]{\delta}$} & $d=2$ & $d=3$ & $d=5$ & $d=10$ & $d=20$ \\ \cline{1-7}
  
  \multirow{3}*{$n=50$ } & 2 clusters & 3.70 & 2.00 & 1.30 & 0.75 & 0.49 \\ \cline{3-7}
  & 3 clusters & 3.10 & 1.80 & 1.20 & 0.700 & 0.55 \\ \cline{3-7}
  & 5 clusters & 2.60 & 1.60 & 1.10 & 0.800 & 0.69 \\ \cline{1-7}
  
  \multirow{3}*{$n=100$} & 2 clusters & 5.20 & 2.50 & 1.50 & 0.80 & 0.53 \\ \cline{3-7}
  & 3 clusters & 4.40 & 2.40 & 1.30 & 0.78 & 0.56 \\ \cline{3-7}
  & 5 clusters & 3.80 & 2.10 & 1.30 & 0.88 & 0.70 \\ \cline{1-7}
  
  \multirow{3}*{$n=200$} & 2 clusters & 7.00 & 3.00 & 1.50 & 0.85 & 0.60 \\ \cline{3-7}
  & 3 clusters & 5.90 & 3.10 & 1.50 & 0.82 & 0.57 \\ \cline{3-7}
  & 5 clusters & 5.40 & 2.70 & 1.50 & 0.98 & 0.74 \\ \cline{1-7}
  
  \multirow{3}*{$n=500$} & 2 clusters & 11.00 & 3.90 & 1.70 & 0.88 & 0.62 \\ \cline{3-7}
  & 3 clusters & 9.00 & 3.50 & 1.90 & 0.85 & 0.60 \\ \cline{3-7}
  & 5 clusters & 8.40 & 3.70 & 1.90 & 1.10 & 0.72 \\ \cline{1-7}

  \end{tabular}
  \caption{The optimal scaled densities ($\sqrt[d]{\delta}$) of KS-CCDs on synthetic datasets consisting of uniform clusters.} \label{tab:uni_KS-CCDs_den_Apdx}
\end{table}

\begin{table}[ht]
  \centering
  \resizebox{\columnwidth}{!}{\begin{tabular}{|c|c|c|c|c|c|c|c|c|c|c|c|c|c|c|c|c|}
  \hline

  \multicolumn{2}{|c|}{\multirow{2}*{KS-CCDs}} & \multicolumn{3}{|c|}{$d=2$} & \multicolumn{3}{|c|}{$d=3$} & \multicolumn{3}{|c|}{$d=5$} & \multicolumn{3}{|c|}{$d=10$} & \multicolumn{3}{|c|}{$d=20$} \\ \cline{3-17}
  \multicolumn{2}{|c|}{} & \textbf{ARI} & \textbf{Sil} & \textbf{SR} & \textbf{ARI} & \textbf{Sil} & \textbf{SR} & \textbf{ARI} & \textbf{Sil} & \textbf{SR} & \textbf{ARI} & \textbf{Sil} & \textbf{SR} & \textbf{ARI} & \textbf{Sil} & \textbf{SR} \\ \hline
  
  \multirow{3}*{$n=50$ } & 2 clusters & 0.997 & 0.698 & 0.998 & 0.999 & 0.665 & 1.000 & 0.999 & 0.632 & 1.000 & 1.000 & 0.608 & 1.000 & 1.000 & 0.594 & 1.000 \\ \cline{3-17}
  & 3 clusters & 0.984 & 0.680 & 0.964 & 0.996 & 0.654 & 0.998 & 0.999 & 0.627 & 1.000 & 1.000 & 0.600 & 1.000 & 1.000 & 0.584 & 1.000 \\ \cline{3-17}
  & 5 clusters & 0.879 & 0.603 & 0.622 & 0.922 & 0.584 & 0.772 & 0.972 & 0.565 & 0.922 & 0.998 & 0.546 & 0.998 & 1.000 & 0.536 & 1.000 \\ \cline{1-17}
  
  \multirow{3}*{$n=100$} & 2 clusters & 1.000 & 0.698 & 1.000 & 1.000 & 0.665 & 1.000 & 1.000 & 0.634 & 1.000 & 1.000 & 0.606 & 1.000 & 1.000 & 0.591 & 1.000 \\ \cline{3-17}
  & 3 clusters & 0.999 & 0.687 & 0.998 & 0.999 & 0.654 & 1.000 & 1.000 & 0.625 & 1.000 & 1.000 & 0.598 & 1.000 & 1.000 & 0.585 & 1.000 \\ \cline{3-17}
  & 5 clusters & 0.952 & 0.619 & 0.860 & 0.962 & 0.590 & 0.900 & 0.989 & 0.567 & 0.976 & 0.999 & 0.547 & 1.000 & 1.000 & 0.535 & 1.000 \\ \cline{1-17}
  
  \multirow{3}*{$n=200$} & 2 clusters & 1.000 & 0.698 & 1.000 & 1.000 & 0.666 & 1.000 & 1.000 & 0.631 & 1.000 & 1.000 & 0.606 & 1.000 & 1.000 & 0.589 & 1.000 \\ \cline{3-17}
  & 3 clusters & 1.000 & 0.688 & 1.000 & 1.000 & 0.655 & 1.000 & 1.000 & 0.626 & 1.000 & 1.000 & 0.600 & 1.000 & 1.000 & 0.587 & 1.000 \\ \cline{3-17}
  & 5 clusters & 0.983 & 0.625 & 0.966 & 0.990 & 0.596 & 0.980 & 0.996 & 0.570 & 1.000 & 1.000 & 0.548 & 1.000 & 1.000 & 0.535 & 1.000 \\ \cline{1-17}
  
  \multirow{3}*{$n=500$} & 2 clusters & 1.000 & 0.696 & 1.000 & 1.000 & 0.666 & 1.000 & 1.000 & 0.634 & 1.000 & 1.000 & 0.604 & 1.000 & 1.000 & 0.590 & 1.000 \\ \cline{3-17}
  & 3 clusters & 1.000 & 0.687 & 1.000 & 1.000 & 0.656 & 1.000 & 1.000 & 0.627 & 1.000 & 1.000 & 0.599 & 1.000 & 1.000 & 0.587 & 1.000 \\ \cline{3-17}
  & 5 clusters & 0.997 & 0.633 & 0.996 & 0.999 & 0.598 & 1.000 & 0.999 & 0.570 & 1.000 & 1.000 & 0.547 & 1.000 & 1.000 & 0.535 & 1.000 \\ \cline{1-17}
  
  \end{tabular}}
  \caption{The simulation results of KS-CCDs on a set of synthetic datasets consisting of uniform clusters. While KS-CCDs generally provide superior clustering performance, their effectiveness drops significantly when identifying a high number of clusters in a small dataset due to low cluster intensity.} \label{tab:uni_KS-CCDs_Apdx}
\end{table}


\begin{table}[htb]
  \centering
  \resizebox{\columnwidth}{!}{\begin{tabular}{|c|c|c|c|c|c|c|c|c|c|c|c|c|c|c|c|c|}
  \hline

  \multicolumn{2}{|c|}{\multirow{2}*{RK-CCDs}} & \multicolumn{3}{|c|}{$d=2$} & \multicolumn{3}{|c|}{$d=3$} & \multicolumn{3}{|c|}{$d=5$} & \multicolumn{3}{|c|}{$d=10$} & \multicolumn{3}{|c|}{$d=20$} \\ \cline{3-17}
  \multicolumn{2}{|c|}{} & \textit{ARI} & \textit{Sil} & \textit{SR} & \textit{ARI} & \textit{Sil} & \textit{SR} & \textit{ARI} & \textit{Sil} & \textit{SR} & \textit{ARI} & \textit{Sil} & \textit{SR} & \textit{ARI} & \textit{Sil} & \textit{SR} \\ \hline
  
  \multirow{3}*{$n=50$ } & 2 clusters & 0.982 & 0.692 & 0.998 & 0.999 & 0.665 & 1.000 & 0.997 & 0.630 & 0.992 & 0.988 & 0.596 & 0.960 & 0.833 & 0.444 & 0.474 \\ \cline{3-17}
  & 3 clusters & 0.906 & 0.655 & 0.844 & 0.990 & 0.651 & 0.988 & 0.998 & 0.625 & 0.994 & 0.972 & 0.589 & 0.920 & 0.681 & 0.424 & 0.382 \\ \cline{3-17}
  & 5 clusters & 0.517 & 0.484 & 0.082 & 0.810 & 0.556 & 0.424 & 0.967 & 0.563 & 0.908 & 0.994 & 0.544 & 0.968 & 0.801 & 0.473 & 0.410 \\ \cline{1-17}
  
  \multirow{3}*{$n=100$} & 2 clusters & 0.999 & 0.697 & 1.000 & 0.999 & 0.665 & 1.000 & 0.999 & 0.633 & 0.998 & 0.979 & 0.584 & 0.922 & 0.834 & 0.437 & 0.454 \\ \cline{3-17}
  & 3 clusters & 0.992 & 0.684 & 0.988 & 0.999 & 0.654 & 0.998 & 0.996 & 0.624 & 0.990 & 0.951 & 0.576 & 0.852 & 0.684 & 0.433 & 0.302 \\ \cline{3-17}
  & 5 clusters & 0.768 & 0.567 & 0.302 & 0.938 & 0.585 & 0.792 & 0.982 & 0.564 & 0.958 & 0.956 & 0.531 & 0.796 & 0.511 & 0.373 & 0.080 \\ \cline{1-17}
  
  \multirow{3}*{$n=200$} & 2 clusters & 0.999 & 0.698 & 1.000 & 1.000 & 0.666 & 1.000 & 1.000 & 0.631 & 1.000 & 0.972 & 0.576 & 0.892 & 0.758 & 0.384 & 0.288 \\ \cline{3-17}
  & 3 clusters & 1.000 & 0.688 & 1.000 & 0.999 & 0.655 & 1.000 & 1.000 & 0.626 & 1.000 & 0.912 & 0.560 & 0.740 & 0.597 & 0.376 & 0.312 \\ \cline{3-17}
  & 5 clusters & 0.926 & 0.610 & 0.748 & 0.988 & 0.595 & 0.968 & 0.993 & 0.568 & 0.976 & 0.917 & 0.513 & 0.626 & 0.595 & 0.391 & 0.130 \\ \cline{1-17}
  
  \multirow{3}*{$n=500$} & 2 clusters & 1.000 & 0.696 & 1.000 & 1.000 & 0.669 & 1.000 & 1.000 & 0.634 & 1.000 & 0.976 & 0.579 & 0.910 & 0.787 & 0.419 & 0.428 \\ \cline{3-17}
  & 3 clusters & 0.999 & 0.686 & 1.000 & 1.000 & 0.654 & 1.000 & 1.000 & 0.627 & 1.000 & 0.929 & 0.571 & 0.806 & 0.592 & 0.394 & 0.208 \\ \cline{3-17}
  & 5 clusters & 0.997 & 0.632 & 1.000 & 0.999 & 0.600 & 1.000 & 0.999 & 0.570 & 1.000 & 0.865 & 0.498 & 0.494 & 0.439 & 0.344 & 0.082 \\ \cline{1-17}
  
  \end{tabular}}
  \caption{The simulation results of RK-CCDs on a set of synthetic datasets consisting of uniform clusters. The performance of RK-CCDs declines sharply in moderate-dimensional spaces due to the limitation of the Ripley's $K$ function.} \label{tab:uni_RK-CCDs_Apdx}
\end{table}


\begin{table}[ht]
  \centering
  \resizebox{\columnwidth}{!}{\begin{tabular}{|c|c|c|c|c|c|c|c|c|c|c|c|c|c|c|c|c|}
  \hline

  \multicolumn{2}{|c|}{\multirow{2}*{UN-CCDs}} & \multicolumn{3}{|c|}{$d=2$} & \multicolumn{3}{|c|}{$d=3$} & \multicolumn{3}{|c|}{$d=5$} & \multicolumn{3}{|c|}{$d=10$} & \multicolumn{3}{|c|}{$d=20$} \\ \cline{3-17}
  \multicolumn{2}{|c|}{} & \textit{ARI} & \textit{Sil} & \textit{SR} & \textit{ARI} & \textit{Sil} & \textit{SR} & \textit{ARI} & \textit{Sil} & \textit{SR} & \textit{ARI} & \textit{Sil} & \textit{SR} & \textit{ARI} & \textit{Sil} & \textit{SR} \\ \hline
  
  \multirow{3}*{$n=50$ } & 2 clusters & 0.980 & 0.691 & 0.996 & 0.996 & 0.664 & 1.000 & 0.999 & 0.632 & 1.000 & 1.000 & 0.608 & 1.000 & 0.992 & 0.585 & 0.968 \\ \cline{3-17}
  & 3 clusters & 0.926 & 0.658 & 0.902 & 0.981 & 0.649 & 0.972 & 0.999 & 0.626 & 1.000 & 1.000 & 0.600 & 1.000 & 0.991 & 0.580 & 0.972 \\ \cline{3-17}
  & 5 clusters & 0.610 & 0.511 & 0.214 & 0.795 & 0.549 & 0.480 & 0.951 & 0.560 & 0.872 & 0.996 & 0.546 & 0.992 & 0.991 & 0.531 & 0.944 \\ \cline{1-17}
  
  \multirow{3}*{$n=100$} & 2 clusters & 0.992 & 0.695 & 0.998 & 0.999 & 0.665 & 1.000 & 1.000 & 0.634 & 1.000 & 1.000 & 0.606 & 1.000 & 0.957 & 0.545 & 0.840 \\ \cline{3-17}
  & 3 clusters & 0.988 & 0.682 & 0.988 & 0.998 & 0.654 & 1.000 & 0.999 & 0.625 & 0.998 & 0.999 & 0.598 & 0.998 & 0.899 & 0.540 & 0.726 \\ \cline{3-17}
  & 5 clusters & 0.847 & 0.583 & 0.604 & 0.935 & 0.582 & 0.836 & 0.986 & 0.565 & 0.976 & 0.998 & 0.546 & 0.992 & 0.957 & 0.521 & 0.802 \\ \cline{1-17}
  
  \multirow{3}*{$n=200$} & 2 clusters & 0.989 & 0.694 & 0.996 & 0.999 & 0.665 & 1.000 & 1.000 & 0.631 & 1.000 & 1.000 & 0.606 & 1.000 & 0.962 & 0.547 & 0.848 \\ \cline{3-17}
  & 3 clusters & 0.994 & 0.685 & 0.998 & 0.999 & 0.655 & 1.000 & 1.000 & 0.626 & 1.000 & 0.998 & 0.600 & 0.996 & 0.921 & 0.558 & 0.802 \\ \cline{3-17}
  & 5 clusters & 0.937 & 0.608 & 0.858 & 0.982 & 0.592 & 0.968 & 0.996 & 0.560 & 0.996 & 0.998 & 0.547 & 0.990 & 0.750 & 0.460 & 0.284 \\ \cline{1-17}
  
  \multirow{3}*{$n=500$} & 2 clusters & 0.992 & 0.693 & 0.998 & 1.000 & 0.666 & 1.000 & 1.000 & 0.634 & 1.000 & 1.000 & 0.604 & 1.000 & 0.962 & 0.548 & 0.850 \\ \cline{3-17}
  & 3 clusters & 0.991 & 0.682 & 1.000 & 1.000 & 0.655 & 1.000 & 1.000 & 0.627 & 1.000 & 0.997 & 0.597 & 0.990 & 0.851 & 0.525 & 0.610 \\ \cline{3-17}
  & 5 clusters & 0.978 & 0.624 & 0.974 & 0.996 & 0.597 & 0.998 & 0.999 & 0.570 & 1.000 & 0.997 & 0.545 & 0.982 & 0.844 & 0.477 & 0.438 \\ \cline{1-17}
  \end{tabular}}
  \caption{The simulation results of UN-CCDs on a set of synthetic datasets consisting of uniform clusters. UN-CCDs perform comparably to RK-CCDs in low dimensions but are substantially better in higher dimensions.} \label{tab:uni_UN-CCDs_Apdx}
\end{table}


According to the simulation result, 
KS-CCDs consistently deliver best overall performance when the optimal $\delta$ is selected,
yielding \textit{ARI} values and success rates (\textit{SR}) close to 1 across most simulation settings,
resulting in high average \textit{Sil} values (consistently above 0.55).
However, in certain cases when $n=50$,
KS-CCDs exhibit reduced efficiency when dealing with 5 clusters.
For example, 
when $d=2$ and $n=50$,
the success rate (\textit{SR}) declines from 0.998 to 0.622
as the number of clusters increases from 2 to 5,
As shown in the Figure \ref{fig:Uniform_ARI_KSCCDs_Apdx},
the ARI value also decreases when the number of cluster increase.
The degradation in performance can be attributed to a drop in the intensities of individual clusters.
Furthermore, the small proximity of the $4^{th}$ and $5^{th}$ clusters to the existing clusters
increases the likelihood of these clusters being erroneously confused as one.
Fortunately, when $n \geq 100$,
KS-CCDs exhibit substantial improvement in handling 5-cluster datasets.
 
Although RK-CCDs are not as effective as KS-CCDs across low-dimensional settings ($d \leq 5$),
they still maintain \textit{ARI} values and success rates (\textit{SR}) above 0.9 in most scenarios,
indicating solid performance in low-dimensional space.
However, similar to the behavior of KS-CCDs,
RK-CCDs also suffer performance degradation in certain cases when there are 5 clusters.
For instance,
when $d=2$ and $n=50$,
the success rate (\textit{SR}) drops drastically from 0.998 to 0.082
as the number of clusters increases from 2 to 5, 
attributable to the low intensities of individual clusters.
Additionally,
RK-CCDs perform worse in as dimension increases,
due to the limitations of the Ripley's $K$ function when $d$ is large.
for example, when $d=10$, 
the success rates (\textit{SR}) decrease substantially, 
reaching a minimum of 0.494 for 5-cluster datasets when $n=500$.
This deterioration is further exacerbated when $d=20$, 
with \textit{ARI} values falling below 0.8 and success rates (\textit{SR}) smaller than 0.4 in most settings.

Under low-dimensional space ($d \leq 5$),
UN-CCDs demonstrate comparable performance to RK-CCDs, 
achieving \textit{ARI} values and success rates (\textit{SR}) exceeding 0.95 in most settings.
Furthermore, UN-CCDs exhibit substantial improvement over RK-CCDs in moderate-dimensional space.
When $d=10$, both \textit{ARI} and success rate approach 1,
much higher than those achieved by RK-CCDs.
The performance gap between RK-CCDs and UN-CCDs becomes even more pronounced when $d=20$.
For example, 
with 3-cluster datasets where $n=200$,
UN-CCDs yield performance metrics of 0.921 (\textit{ARI}), 0.558 (\textit{Sil}), and 0.802 (success rate), respectively.
These values are substantially higher than the corresponding 0.597, 0.376, and 0.312 obtained by RK-CCDs.
As we talked previously in Section \ref{sec:MC-SRT_NND},
this improvement is attributed to the newly developed MC-SRT,
which employed NND instead of the Ripley's $K$ function with several additional key enhancements.

To explicitly evaluate scalability,
we recorded the total empirical runtime for each CCD variant under the uniform cluster setting at $d=5$,
varying the sample size $n$ from 50 to 500.
Simulations were executed on a Linux workstation equipped with an Intel Core i9-13900K processor and 6400MHz 64 GB DDR5 RAM,
with multithreaded parallel computing enabled.
The runtimes reported in Table \ref{tab:runtime_comparison} reflect the total duration in minutes required to generate the random datasets with number of clusters varying from 2 to 5,
execute the clustering algorithms,
and complete 500 independent repetitions for average performance.

UN-CCDs demonstrate substantial empirical efficiency,
completing the entire 500-iteration process for $n=500$ in just 4.38 minutes.
In contrast, the parameter-free baseline RK-CCDs required 78.60 minutes under identical conditions.
KS-CCDs incurred a severely inflated computational cost---exceeding 980 minutes at $n=500$---because optimizing the density parameter $\delta$ required an exhaustive grid search over 200 candidate values ranging from 0.01 to 20 for each generated dataset.
This runtime comparison confirms that, within its intended practical regime of moderate sample sizes,
UN-CCDs are empirically efficient relative to the other CCD variants while retaining the practical advantage of requiring little parameter tuning.

\begin{table}[ht]
\centering
\caption{Total empirical runtime (in minutes) for 500 simulation iterations at $d=5$.}
\label{tab:runtime_comparison}
\begin{tabular}{lrrrr}
\hline
Method & $n=50$ & $n=100$ & $n=200$ & $n=500$ \\
\hline
UN-CCDs & 0.15 & 0.29 & 0.65 & 4.38 \\
RK-CCDs & 4.60 & 6.46 & 16.99 & 78.60 \\
KS-CCDs & 38.55 & 41.30 & 85.80 & 986.45 \\
\hline
\end{tabular}
\end{table}

\subsection{Experiment with Gaussian Settings}

\begin{figure}[htb]
    \centering
    \begin{subfigure}{0.32\linewidth}
        \includegraphics[width=\linewidth]{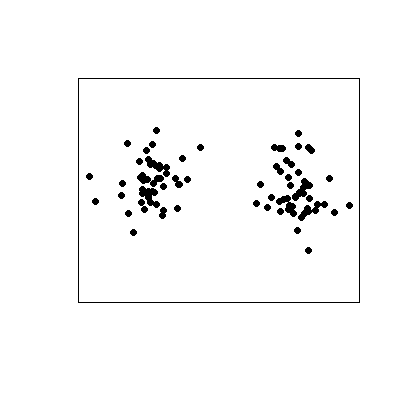}
        \caption{2 clusters}
        \label{fig:gauss_2cls_Apdx} 
    \end{subfigure}
    \hfill 
    \begin{subfigure}{0.32\linewidth}
        \includegraphics[width=\linewidth]{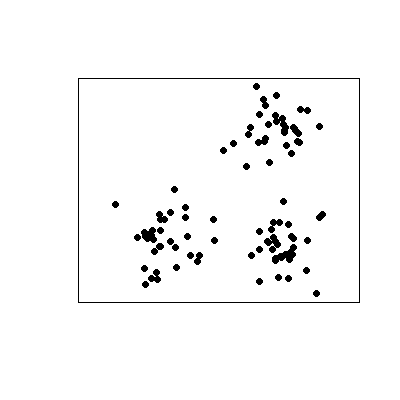}
        \caption{3 clusters}
        \label{fig:gauss_3cls_Apdx} 
    \end{subfigure}
    \hfill 
    \begin{subfigure}{0.32\linewidth}
        \includegraphics[width=\linewidth]{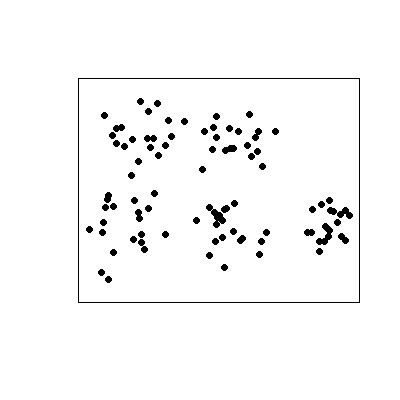}
        \caption{5 clusters}
        \label{fig:gauss_5cls_Apdx} 
    \end{subfigure}
    \caption{Realizations of the simulation settings with 2, 3, and 5 Gaussian clusters in $\mathbb{R}^2$.}
    \label{fig:Demo_2d_Gaussian_Cls_Apdx}
\end{figure}

\begin{figure}[htb]
    \centering
    \begin{subfigure}{1\linewidth}
        \centering
        \includegraphics[width=\linewidth]{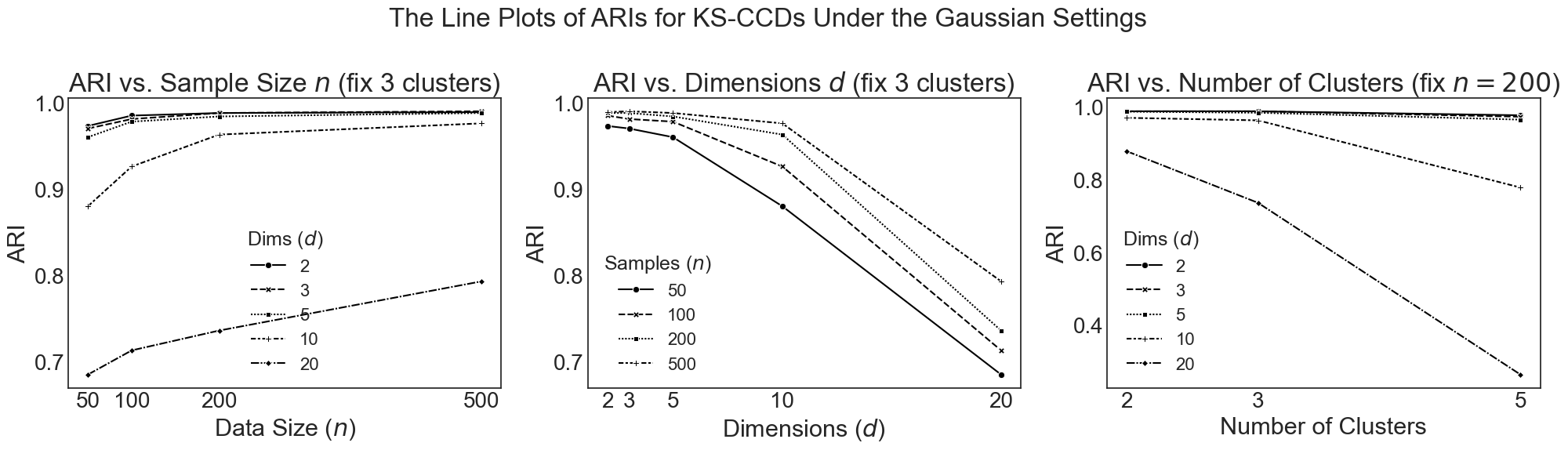}
    \end{subfigure}
    \caption{The line plots of the ARIs of KS-CCDs, under the Gaussian cluster settings.}
    \label{fig:Gaussian_ARI_KSCCDs_Apdx}
\end{figure}

\begin{figure}[htb]
    \centering
    \begin{subfigure}{1\linewidth}
        \centering
        \includegraphics[width=\linewidth]{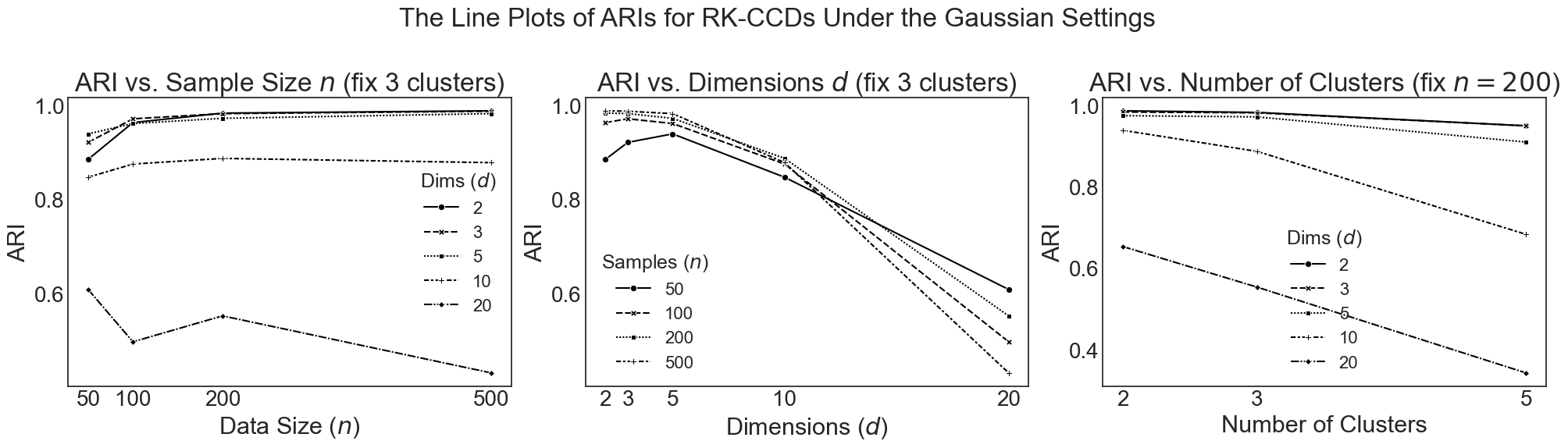}
    \end{subfigure}
    \caption{The line plots of the ARIs of RK-CCDs, under the Gaussian cluster settings.}
    \label{fig:Gaussian_ARI_RKCCDs_Apdx}
\end{figure}

\begin{figure}[htb]
    \centering
    \begin{subfigure}{1\linewidth}
        \centering
        \includegraphics[width=\linewidth]{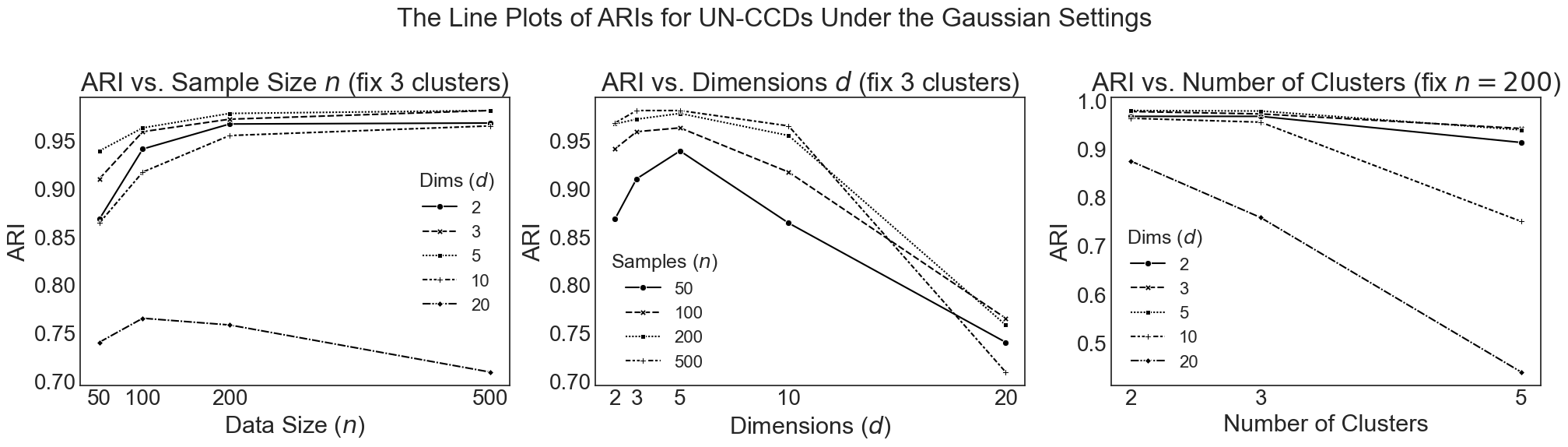}
    \end{subfigure}
    \caption{The line plots of the ARIs of UN-CCDs, under the Gaussian cluster settings.}
    \label{fig:Gaussian_ARI_UNCCDs_Apdx}
\end{figure}

Assuming CSR,
UN-CCDs and RK-CCDs perform well with uniform clusters.
To evaluate the performance of CCD-based clustering methods on non-uniform data,
we conduct simulations with Gaussian clusters.
Each Gaussian cluster has a covariance matrix $\Delta * I_d$,
where $\Delta$ is a random uniform variable between 0.8 and 1.2 (for random scaling),
and $I_d$ represents a $d \times d$ identity matrix.
All other settings, 
including dimensionality,
data size, 
cluster centers, 
and the number of clusters, 
remain the same as in the uniform settings.
Three realizations with different number of Gaussian clusters ($d=2$) are presented in Figure \ref{fig:Demo_2d_Gaussian_Cls_Apdx}.
The simulation results are summarized in Table \ref{tab:Gau_KS-CCDs_Apdx}, \ref{tab:Gau_RK-CCDs_Apdx}, and \ref{tab:Gau_UN-CCDs_Apdx}.
Similarly, we present the performance trend (ARIs) of the CCD-based methods from Figures \ref{fig:Gaussian_ARI_KSCCDs_Apdx} to \ref{fig:Gaussian_ARI_UNCCDs_Apdx}. 


\begin{table}[htb]
  \centering
  \begin{tabular}{|c|c|c|c|c|c|c|}
  \hline

  \multicolumn{2}{|c|}{Density Parameter: $\sqrt[d]{\delta}$} & $d=2$ & $d=3$ & $d=5$ & $d=10$ & $d=20$ \\ \cline{1-7}
  
  \multirow{3}*{$n=50$ } & 2 clusters & 1.90 & 1.10 & 0.60 & 0.32 & 0.20 \\ \cline{3-7}
  & 3 clusters & 1.60 & 0.93 & 0.52 & 0.30 & 0.19 \\ \cline{3-7}
  & 5 clusters & 1.30 & 0.79 & 0.46 & 0.28 & 0.18 \\ \cline{1-7}
  
  \multirow{3}*{$n=100$} & 2 clusters & 3.00 & 1.40 & 0.71 & 0.35 & 0.21 \\ \cline{3-7}
  & 3 clusters & 2.30 & 1.20 & 0.65 & 0.32 & 0.20 \\ \cline{3-7}
  & 5 clusters & 1.90 & 0.99 & 0.58 & 0.32 & 0.17 \\ \cline{1-7}
  
  \multirow{3}*{$n=200$} & 2 clusters & 4.20 & 1.70 & 0.76 & 0.39 & 0.23 \\ \cline{3-7}
  & 3 clusters & 3.50 & 1.50 & 0.71 & 0.37 & 0.21 \\ \cline{3-7}
  & 5 clusters & 2.70 & 1.30 & 0.66 & 0.35 & 0.17 \\ \cline{1-7}
  
  \multirow{3}*{$n=500$} & 2 clusters & 7.1 & 2.5 & 0.95 & 0.43 & 0.24 \\ \cline{3-7}
  & 3 clusters & 5.6 & 2.1 & 0.90 & 0.41 & 0.24 \\ \cline{3-7}
  & 5 clusters & 4.3 & 1.8 & 0.82 & 0.41 & 0.19 \\ \cline{1-7}
  
  \end{tabular}
  \caption{The optimal scaled densities ($\sqrt[d]{\delta}$) of KS-CCDs on synthetic datasets consisting of Gaussian clusters.} \label{tab:Gau_KS-CCDs_den_Apdx}
\end{table}

\begin{table}[htb]
  \centering
  \resizebox{\columnwidth}{!}{\begin{tabular}{|c|c|c|c|c|c|c|c|c|c|c|c|c|c|c|c|c|}
  \hline

  \multicolumn{2}{|c|}{\multirow{2}*{KS-CCDs}} & \multicolumn{3}{|c|}{$d=2$} & \multicolumn{3}{|c|}{$d=3$} & \multicolumn{3}{|c|}{$d=5$} & \multicolumn{3}{|c|}{$d=10$} & \multicolumn{3}{|c|}{$d=20$} \\ \cline{3-17}
  \multicolumn{2}{|c|}{} & \textit{ARI} & \textit{Sil} & \textit{SR} & \textit{ARI} & \textit{Sil} & \textit{SR} & \textit{ARI} & \textit{Sil} & \textit{SR} & \textit{ARI} & \textit{Sil} & \textit{SR} & \textit{ARI} & \textit{Sil} & \textit{SR} \\ \hline
  
  \multirow{3}*{$n=50$ } & 2 clusters & 0.983 & 0.704 & 1.000 & 0.980 & 0.633 & 1.000 & 0.974 & 0.537 & 1.000 & 0.947 & 0.399 & 1.000 & 0.798 & 0.247 & 0.960 \\ \cline{3-17}
  & 3 clusters & 0.973 & 0.688 & 0.984 & 0.970 & 0.619 & 0.986 & 0.960 & 0.524 & 0.970 & 0.880 & 0.381 & 0.832 & 0.685 & 0.251 & 0.470 \\ \cline{3-17}
  & 5 clusters & 0.936 & 0.644 & 0.890 & 0.921 & 0.569 & 0.872 & 0.865 & 0.470 & 0.718 & 0.651 & 0.338 & 0.376 & 0.380 & 0.244 & 0.050 \\ \cline{1-17}
  
  \multirow{3}*{$n=100$} & 2 clusters & 0.980 & 0.704 & 1.000 & 0.987 & 0.635 & 1.000 & 0.982 & 0.540 & 1.000 & 0.960 & 0.399 & 0.998 & 0.846 & 0.256 & 0.980 \\ \cline{3-17}
  & 3 clusters & 0.985 & 0.693 & 1.000 & 0.981 & 0.621 & 0.996 & 0.978 & 0.527 & 0.996 & 0.926 & 0.383 & 0.920 & 0.713 & 0.254 & 0.508 \\ \cline{3-17}
  & 5 clusters & 0.963 & 0.652 & 0.986 & 0.962 & 0.582 & 0.982 & 0.939 & 0.475 & 0.936 & 0.717 & 0.341 & 0.556 & 0.303 & 0.249 & 0.000 \\ \cline{1-17}
  
  \multirow{3}*{$n=200$} & 2 clusters & 0.988 & 0.703 & 1.000 & 0.988 & 0.634 & 1.000 & 0.987 & 0.539 & 1.000 & 0.970 & 0.402 & 1.000 & 0.878 & 0.261 & 0.982 \\ \cline{3-17}
  & 3 clusters & 0.988 & 0.694 & 1.000 & 0.988 & 0.623 & 1.000 & 0.984 & 0.527 & 1.000 & 0.963 & 0.389 & 0.988 & 0.736 & 0.257 & 0.550 \\ \cline{3-17}
  & 5 clusters & 0.977 & 0.657 & 0.998 & 0.973 & 0.581 & 0.998 & 0.965 & 0.481 & 0.988 & 0.778 & 0.342 & 0.704 & 0.263 & 0.254 & 0.000 \\ \cline{1-17}
  
  \multirow{3}*{$n=500$} & 2 clusters & 0.990 & 0.703 & 1.000 & 0.991 & 0.637 & 1.000 & 0.989 & 0.541 & 1.000 & 0.978 & 0.404 & 1.000 & 0.905 & 0.266 & 0.998 \\ \cline{3-17}
  & 3 clusters & 0.989 & 0.691 & 1.000 & 0.990 & 0.625 & 1.000 & 0.988 & 0.528 & 1.000 & 0.976 & 0.391 & 0.996 & 0.793 & 0.258 & 0.652 \\ \cline{3-17}
  & 5 clusters & 0.981 & 0.658 & 1.000 & 0.980 & 0.586 & 1.000 & 0.974 & 0.483 & 1.000 & 0.824 & 0.347 & 0.776 & 0.297 & 0.255 & 0.000 \\ \cline{1-17}
  
  \end{tabular}}
  \caption{The simulation results of KS-CCDs on a set of synthetic datasets consisting of Gaussian clusters. While highly efficient with Gaussian clusters in low dimensions, KS-CCDs' performance degrades substantially in higher dimensional spaces due to the decreased intensity and unbounded nature of Gaussian distributions.} \label{tab:Gau_KS-CCDs_Apdx}
\end{table}


\begin{table}[htb]
  \centering
  \resizebox{\columnwidth}{!}{\begin{tabular}{|c|c|c|c|c|c|c|c|c|c|c|c|c|c|c|c|c|}
  \hline

  \multicolumn{2}{|c|}{\multirow{2}*{KS-CCDs}} & \multicolumn{3}{|c|}{$d=2$} & \multicolumn{3}{|c|}{$d=3$} & \multicolumn{3}{|c|}{$d=5$} & \multicolumn{3}{|c|}{$d=10$} & \multicolumn{3}{|c|}{$d=20$} \\ \cline{3-17}
  \multicolumn{2}{|c|}{} & \textit{ARI} & \textit{Sil} & \textit{SR} & \textit{ARI} & \textit{Sil} & \textit{SR} & \textit{ARI} & \textit{Sil} & \textit{SR} & \textit{ARI} & \textit{Sil} & \textit{SR} & \textit{ARI} & \textit{Sil} & \textit{SR} \\ \hline
  
  \multirow{3}*{$n=50$ } & 2 clusters & 0.950 & 0.691 & 1.000 & 0.964 & 0.627 & 1.000 & 0.949 & 0.528 & 0.994 & 0.908 & 0.384 & 0.978 & 0.660 & 0.181 & 0.628 \\ \cline{3-17}
  & 3 clusters & 0.885 & 0.659 & 0.812 & 0.921 & 0.607 & 0.882 & 0.939 & 0.517 & 0.944 & 0.847 & 0.372 & 0.784 & 0.609 & 0.212 & 0.304 \\ \cline{3-17}
  & 5 clusters & 0.627 & 0.535 & 0.232 & 0.760 & 0.521 & 0.448 & 0.812 & 0.453 & 0.646 & 0.654 & 0.328 & 0.452 & 0.396 & 0.222 & 0.078 \\ \cline{1-17}
  
  \multirow{3}*{$n=100$} & 2 clusters & 0.982 & 0.702 & 1.000 & 0.980 & 0.632 & 1.000 & 0.972 & 0.536 & 0.998 & 0.917 & 0.381 & 0.958 & 0.664 & 0.181 & 0.630 \\ \cline{3-17}
  & 3 clusters & 0.963 & 0.685 & 0.962 & 0.971 & 0.617 & 0.984 & 0.961 & 0.521 & 0.978 & 0.875 & 0.370 & 0.814 & 0.498 & 0.186 & 0.218 \\ \cline{3-17}
  & 5 clusters & 0.830 & 0.600 & 0.594 & 0.906 & 0.561 & 0.798 & 0.874 & 0.459 & 0.802 & 0.664 & 0.327 & 0.462 & 0.348 & 0.199 & 0.046 \\ \cline{1-17}
  
  \multirow{3}*{$n=200$} & 2 clusters & 0.987 & 0.702 & 1.000 & 0.984 & 0.633 & 1.000 & 0.975 & 0.535 & 1.000 & 0.939 & 0.387 & 0.954 & 0.653 & 0.170 & 0.548 \\ \cline{3-17}
  & 3 clusters & 0.983 & 0.692 & 0.998 & 0.982 & 0.621 & 0.998 & 0.972 & 0.522 & 0.988 & 0.887 & 0.375 & 0.826 & 0.553 & 0.188 & 0.272 \\ \cline{3-17}
  & 5 clusters & 0.950 & 0.646 & 0.934 & 0.950 & 0.572 & 0.932 & 0.910 & 0.467 & 0.870 & 0.683 & 0.328 & 0.500 & 0.342 & 0.201 & 0.048 \\ \cline{1-17}
  
  \multirow{3}*{$n=500$} & 2 clusters & 0.989 & 0.702 & 1.000 & 0.990 & 0.633 & 1.000 & 0.984 & 0.540 & 1.000 & 0.939 & 0.383 & 0.912 & 0.578 & 0.152 & 0.500 \\ \cline{3-17}
  & 3 clusters & 0.988 & 0.690 & 1.000 & 0.987 & 0.624 & 1.000 & 0.982 & 0.525 & 0.996 & 0.878 & 0.370 & 0.780 & 0.432 & 0.165 & 0.174 \\ \cline{3-17}
  & 5 clusters & 0.977 & 0.655 & 0.994 & 0.976 & 0.581 & 1.000 & 0.951 & 0.475 & 0.960 & 0.668 & 0.328 & 0.454 & 0.335 & 0.184 & 0.066 \\ \cline{1-17}
  
  \end{tabular}}
  \caption{The simulation results of RK-CCDs on a set of synthetic datasets consisting of Gaussian clusters. Similar to Uniform settings, the performance of RK-CCDs deteriorates significantly in higher dimensional space due to the limitations of the Ripley's $K$ function.} \label{tab:Gau_RK-CCDs_Apdx}
\end{table}


\begin{table}[htb]
  \centering
  \resizebox{\columnwidth}{!}{\begin{tabular}{|c|c|c|c|c|c|c|c|c|c|c|c|c|c|c|c|c|}
  \hline

  \multicolumn{2}{|c|}{\multirow{2}*{UN-CCDs}} & \multicolumn{3}{|c|}{$d=2$} & \multicolumn{3}{|c|}{$d=3$} & \multicolumn{3}{|c|}{$d=5$} & \multicolumn{3}{|c|}{$d=10$} & \multicolumn{3}{|c|}{$d=20$} \\ \cline{3-17}
  \multicolumn{2}{|c|}{} & \textit{ARI} & \textit{Sil} & \textit{SR} & \textit{ARI} & \textit{Sil} & \textit{SR} & \textit{ARI} & \textit{Sil} & \textit{SR} & \textit{ARI} & \textit{Sil} & \textit{SR} & \textit{ARI} & \textit{Sil} & \textit{SR} \\ \hline
  
  \multirow{3}*{$n=50$ } & 2 clusters & 0.930 & 0.683 & 0.988 & 0.958 & 0.625 & 1.000 & 0.953 & 0.530 & 0.998 & 0.943 & 0.395 & 1.000 & 0.850 & 0.249 & 0.988 \\ \cline{3-17}
  & 3 clusters & 0.868 & 0.649 & 0.854 & 0.910 & 0.600 & 0.912 & 0.939 & 0.517 & 0.944 & 0.864 & 0.374 & 0.820 & 0.740 & 0.248 & 0.574 \\ \cline{3-17}
  & 5 clusters & 0.644 & 0.524 & 0.346 & 0.740 & 0.515 & 0.470 & 0.774 & 0.446 & 0.578 & 0.600 & 0.330 & 0.320 & 0.444 & 0.232 & 0.136 \\ \cline{1-17}
  
  \multirow{3}*{$n=100$} & 2 clusters & 0.960 & 0.695 & 0.994 & 0.971 & 0.630 & 0.998 & 0.974 & 0.537 & 1.000 & 0.951 & 0.394 & 1.000 & 0.859 & 0.248 & 0.962 \\ \cline{3-17}
  & 3 clusters & 0.941 & 0.676 & 0.958 & 0.959 & 0.612 & 0.978 & 0.963 & 0.522 & 0.984 & 0.917 & 0.379 & 0.904 & 0.765 & 0.247 & 0.618 \\ \cline{3-17}
  & 5 clusters & 0.828 & 0.593 & 0.670 & 0.888 & 0.555 & 0.816 & 0.875 & 0.462 & 0.810 & 0.713 & 0.336 & 0.554 & 0.438 & 0.232 & 0.128 \\ \cline{1-17}
  
  \multirow{3}*{$n=200$} & 2 clusters & 0.967 & 0.696 & 1.000 & 0.977 & 0.631 & 1.000 & 0.979 & 0.537 & 1.000 & 0.963 & 0.399 & 1.000 & 0.875 & 0.254 & 0.978 \\ \cline{3-17}
  & 3 clusters & 0.967 & 0.685 & 0.996 & 0.972 & 0.618 & 0.996 & 0.978 & 0.524 & 1.000 & 0.955 & 0.386 & 0.972 & 0.758 & 0.250 & 0.586 \\ \cline{3-17}
  & 5 clusters & 0.913 & 0.628 & 0.902 & 0.942 & 0.569 & 0.950 & 0.939 & 0.473 & 0.952 & 0.750 & 0.337 & 0.648 & 0.439 & 0.235 & 0.134 \\ \cline{1-17}
  
  \multirow{3}*{$n=500$} & 2 clusters & 0.971 & 0.697 & 1.000 & 0.982 & 0.635 & 1.000 & 0.983 & 0.540 & 1.000 & 0.973 & 0.402 & 1.000 & 0.885 & 0.253 & 0.946 \\ \cline{3-17}
  & 3 clusters & 0.968 & 0.683 & 0.996 & 0.981 & 0.621 & 1.000 & 0.981 & 0.525 & 1.000 & 0.965 & 0.387 & 0.988 & 0.799 & 0.249 & 0.662 \\ \cline{3-17}
  & 5 clusters & 0.950 & 0.643 & 0.984 & 0.963 & 0.578 & 0.994 & 0.963 & 0.477 & 0.998 & 0.808 & 0.342 & 0.760 & 0.450 & 0.232 & 0.140 \\ \cline{1-17}
  
  \end{tabular}}
  \caption{The simulation results of UN-CCDs on a set of synthetic datasets consisting of Gaussian clusters. The UN-CCD method demonstrate substantial advantage over KS-CCDs and RK-CCDs in higher dimensional space.} \label{tab:Gau_UN-CCDs_Apdx}
\end{table}


While the optimal density parameter $\delta$ is selected for each simulation setting,
KS-CCDs demonstrate reduced effectiveness with Gaussian clusters.
In low-dimensional settings ($d \leq 5$), 
KS-CCDs deliver slightly lower performance in the comparable scenarios with uniform clusters.
Notably, \textit{ARI} values and success rates (\textit{SR}) remain high ($>0.95$), 
indicating high efficiency even with Gaussian clusters.
However,
as shown in the second line plot of Figure \ref{fig:Gaussian_ARI_KSCCDs_Apdx}.
KS-CCDs degrade substantially in moderate-dimensional settings ($d \geq 10$),
particularly with three or more clusters.
For example, 
when $d=20$ and $n=200$,
KS-CCDs deliver average performance metrics of 0.736 (\textit{ARI}), 0.257 (\textit{Sil}), and 0.550 (success rate) for 3-cluster datasets,
drastically lower than the corresponding 1, 0.587, and 1 obtained with uniform clusters.
This performance decline can be explained by the decreased intensities of Gaussian clusters in moderate-dimensional space,
an effect amplified with a larger number of clusters (e.g., 5 clusters),
suggested by the third line plot of Figure \ref{fig:Gaussian_ARI_KSCCDs_Apdx}.
Moreover, the support of Gaussian clusters is unbounded,
further contributes to the performance degradation.

Similar to the simulations with uniform clusters, 
RK-CCDs generally perform well in low-dimensional space,
with \textit{ARI} values and success rates (\textit{SR}) exceeding 0.9
except certain cases with 5 clusters,
where the proximities between clusters and the intensities of individual clusters are small.
For instance,
with $d=2$, $n=50$, and 5 clusters,
the performance metrics of RK-CCDs are 0.627 (\textit{ARI}), 0.535 (\textit{Sil}), and 0.234 (success rate), respectively.
RK-CCDs demonstrate similar behavior in moderate-dimensional space ($d \geq 10$) as in the uniform settings.
Their performance is slightly worse than KS-CCDs when $d=10$,
and deteriorate further when $d=20$,
increasing the data size does not help.
For example,
when $d=20$ and $n=500$,
the \textit{ARI} and success rate are 0.297 and 0, respectively,
due to the limitations of the MC-SRT based on the Ripley's \textit{K} function.

Similar to the previous analysis, 
UN-CCDs perform similarly to RK-CCDs and KS-CCDs in low-dimensional space ($ d \leq 5$),
Both \textit{ARI} values and success rates exceed 0.9 in most scenarios, 
with a few exceptions observed when dealing with 5 clusters. 
These results are slightly lower than those obtained with uniform clusters.
However,
UN-CCDs yield robust and completable overall performance in higher dimensional space ($d \geq 10$).
When $d = 10$,
UN-CCDs deliver comparable performance to KS-CCDs,
outperforming RK-CCDs,
as evidenced by higher \textit{ARI} values and success rates.
To illustrate,
with $n=200$, $d=10$, and 3 clusters,
the \textit{ARI} values for KS-CCDs, RK-CCDs, and UN-CCDs are 0.963, 0.877, and 0.955, respectively.
Furthermore,
UN-CCDs exhibit a substantial advantage over the other two CCDs when $d=20$.
For example,
with $n=200$, $d=20$, and 5 clusters,
the \textit{ARI} values for KS-CCDs, RK-CCDs, and UN-CCDs are 0.263, 0.342, and 0.439, respectively.

Overall,
when the optimal density parameter $\delta$ is selected, 
KS-CCDs yielded the most accurate partitions for uniform clusters across all dimensions, 
consistently achieving \textit{ARI} values and success rates near 1 in most scenarios, 
invariant to the varying dimensions $d$.
Meanwhile, among the parameter-free approaches,
UN-CCDs showed a clear advantage as data complexity increased.
While both RK-CCDs and UN-CCDs performed well in lower dimensions ($d \leq 5$),
RK-CCDs degraded sharply at $d \geq 10$.
In contrast, UN-CCDs maintained robust performance in these moderate dimensions.
When with uniform clusters,
UN-CCDs exhibit slightly lower performance compared to KS-CCDs,
and show a substantial advantage over RK-CCDs when $d \geq 10$.
This advantage became even more pronounced with Gaussian clusters, 
where UN-CCDs outperformed both baseline methods in moderate dimensions at $d \geq 10$.
Finally, we observe that the performance of all three CCDs declines with 5 clusters,
particularly when the data size $n$ is small,
primarily because the sparse sampling fails to provide sufficient intensity to define clear cluster boundaries.

\subsection{Experiment on Datasets with Noise}
\label{sec:simulation_W_noise}
\begin{figure}[htb]
    \centering
    \begin{subfigure}{0.40\textwidth}
        \includegraphics[width=\linewidth]{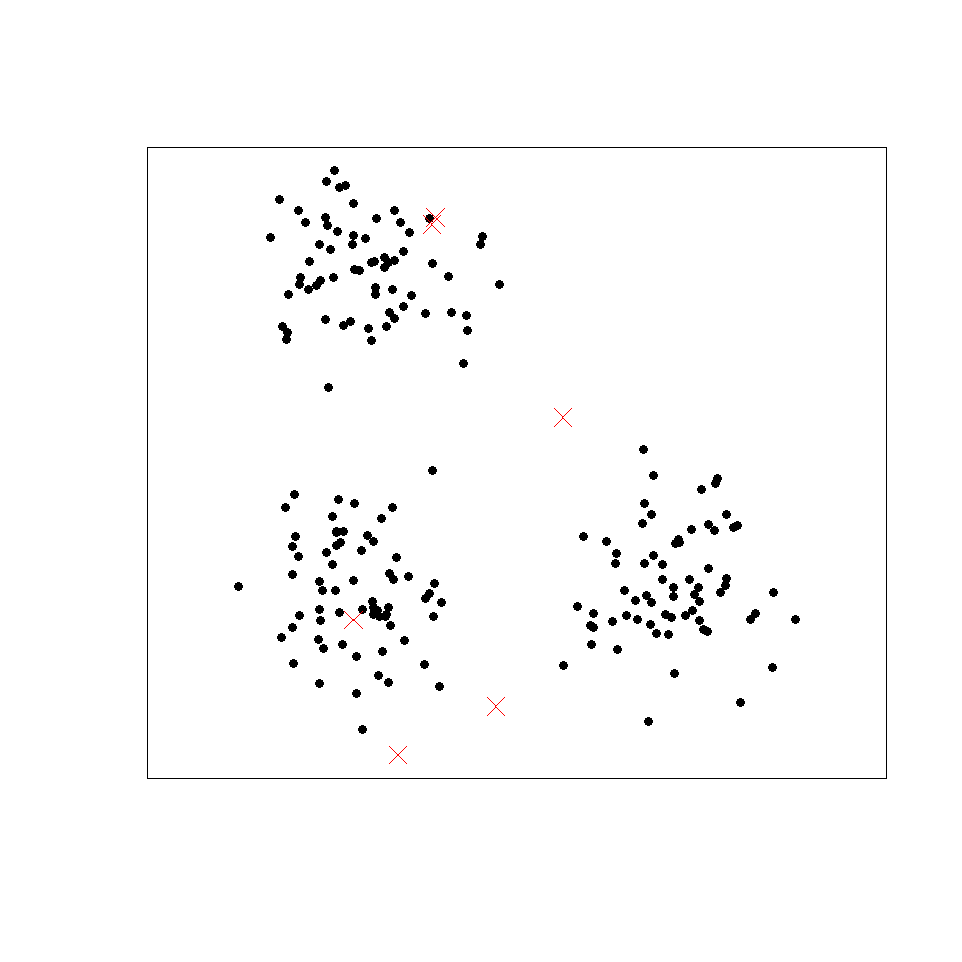}
        \caption{3\%}
        \label{fig:noise_3_Apdx} 
    \end{subfigure}
    \hfill 
    \begin{subfigure}{0.40\textwidth}
        \includegraphics[width=\linewidth]{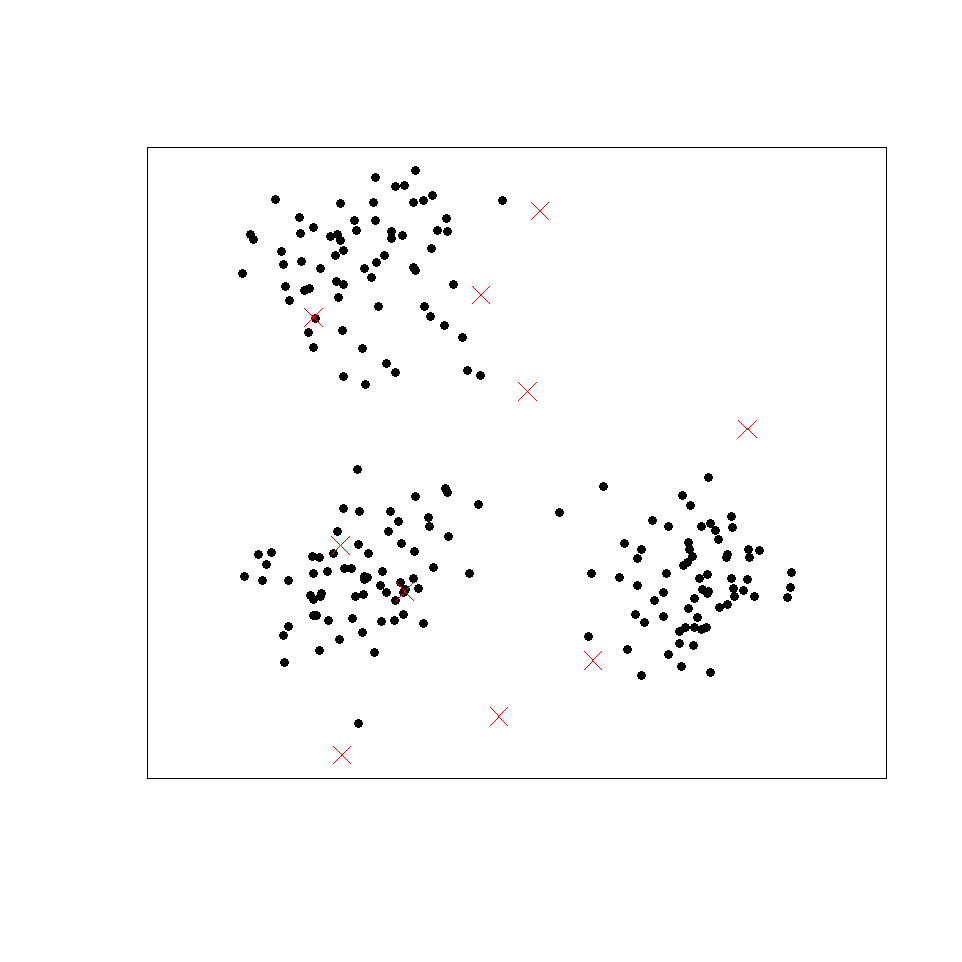}
        \caption{5\%}
        \label{fig:noise_5_Apdx} 
    \end{subfigure}

    
    \begin{subfigure}{0.40\textwidth}
        \includegraphics[width=\linewidth]{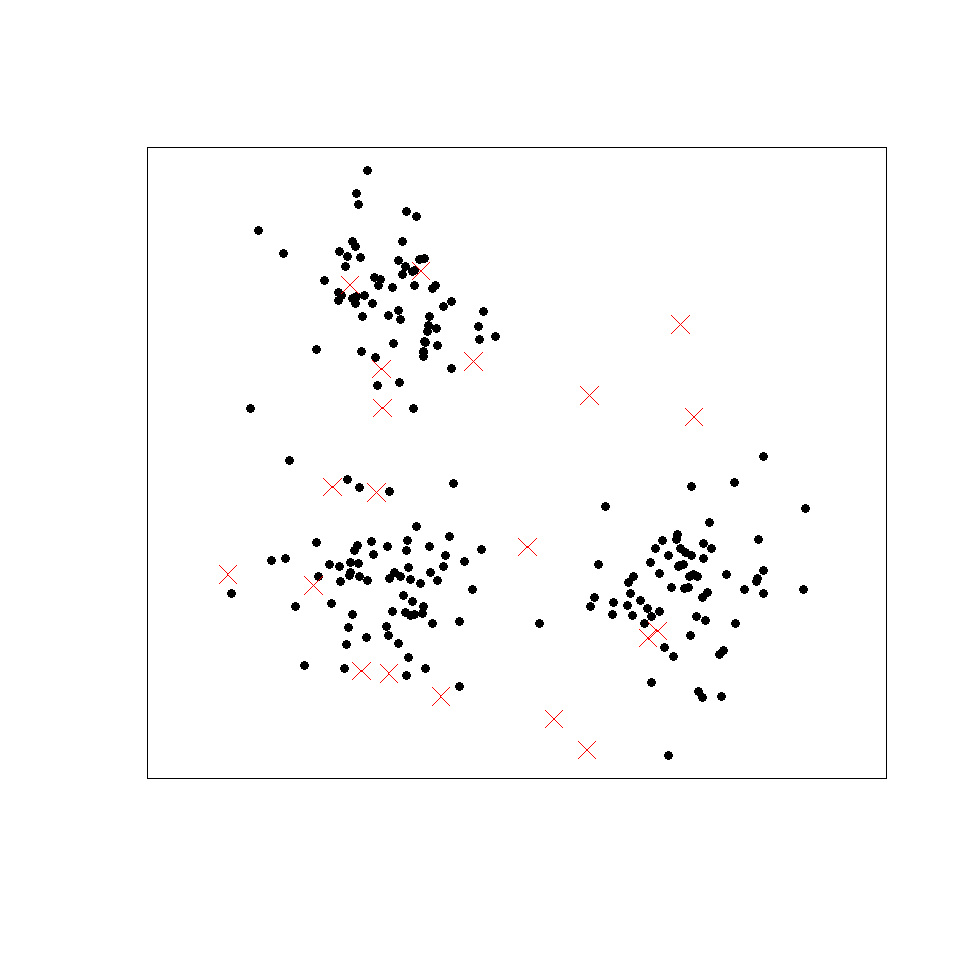}
        \caption{10\%}
        \label{fig:noise_10_Apdx} 
    \end{subfigure}
    \hfill 
    \begin{subfigure}{0.40\textwidth}
        \includegraphics[width=\linewidth]{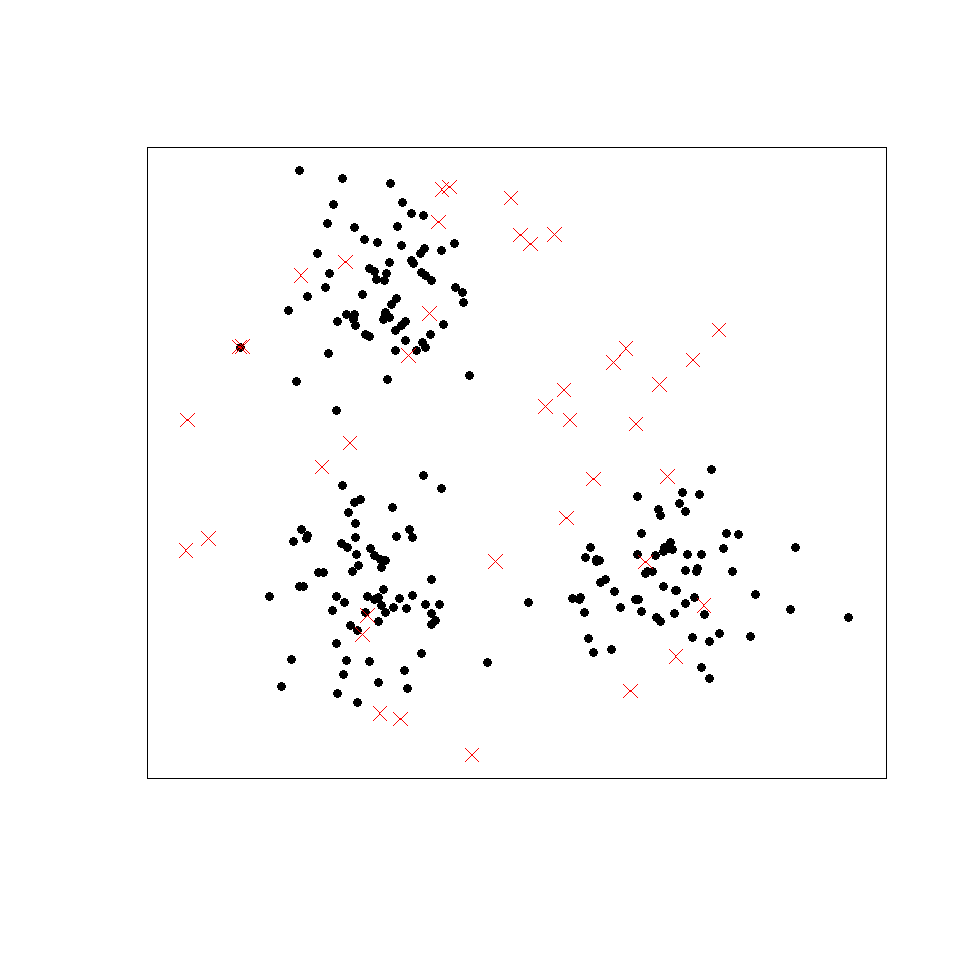}
        \caption{20\%}
        \label{fig:noise_20_Apdx} 
    \end{subfigure}
    
    \caption{Some realizations of the simulation studying the impact of different noise levels, the noise level varies from $3\%$ to $20\%$. Red crosses are noise, black points are regular observations. The noise levels are indicated below each sub-figure.}
    \label{fig:Demo_Noise_Level_Clustering_Apdx}
\end{figure}

In this section,
we conduct additional simulation experiments to evaluate whether the UN-CCD method is robust to noise.
In the context of this study, 
we distinguish noise from general outlier detection by defining it specifically as randomly distributed background points.
The general simulation setup is similar to the experiments with Gaussian settings,
where each Gaussian cluster has a random scaling covariance matrix $\Delta * I_d$.
To simplify the study,
we set the number of clusters to 3,
and the data size to 200 plus the number of noise points.
Noise refers to the points that are randomly distributed within the scope of the entire dataset.
we simulate datasets with varying noise level ($3\%$, $5\%$, $10\%$, $20\%$) to assess the performance trend of UN-CCD methods.
Additionally,
we consider both 3-dimensional and 10-dimensional datasets to study the impact of background noise level with different dimensions.
Specific details are outlined below.
Some realizations of datasets with Gaussian clusters in 2-dimensional space (although the simulation experiments are conducted on 3 and 10-dimensional space) are presented in Figure \ref{fig:Demo_Noise_Level_Clustering_Apdx} (for illustration purposes).
Similar to the previous simulation study, 
Each simulation setting is repeated 1000 times to ensure accurate results.
The simulation results are summarized in Table \ref{tab:Noise_Level_Clustering_Apdx},
similar to the previous simulations,
we record the average \textit{ARI}, \textit{Sil}, and \textit{SR} under each setting.

\begin{itemize}
  \item[\romannumeral1.] The dimensionality ($d$) of the simulated datasets: 3, 10;
  \item[\romannumeral2.] The noise levels ($l$): $3\%$, $5\%$, $10\%$, $20\%$
  \item[\romannumeral3.] The size of datasets ($n$): $200 \times (1 + l)$;
  \item[\romannumeral4.] The size of each cluster is equal;
  \item[\romannumeral5.] The number of clusters: 3;
  \item[\romannumeral6.] The centers of Gaussian clusters: $\bm{\mu_1} = (\underbrace{3,...,3}_{d})$, $\bm{\mu_2} = (9,\underbrace{3,...,3}_{d-1})$, and $\bm{\mu_3} = (3,9,\underbrace{3,...,3}_{d-2})$.
  \item[\romannumeral7.] The covariance matrix of each Gaussian cluster: $\Delta * I_d$, where $\Delta \sim U[0.8, 1.2]$ and $I_d$ is a $d$-dimensional identity matrix.
  \label{sec:Noise_Setting_Clustering_Apdx}
\end{itemize}


\begin{table}[htb]
  \centering
  \resizebox{\columnwidth}{!}{\begin{tabular}{|c|c|c|c|c|c|c|c|c|c|c|c|c|c|}
  \hline

  \multicolumn{2}{|c|}{\multirow{2}*{UN-CCDs}} & \multicolumn{3}{|c|}{$l = 3\%$} & \multicolumn{3}{|c|}{$l = 5\%$} & \multicolumn{3}{|c|}{$l = 10\%$} & \multicolumn{3}{|c|}{$l = 20\%$} \\ \cline{3-14}
  \multicolumn{2}{|c|}{} & \textit{ARI} & \textit{Sil} & \textit{SR} & \textit{ARI} & \textit{Sil} & \textit{SR} & \textit{ARI} & \textit{Sil} & \textit{SR} & \textit{ARI} & \textit{Sil} & \textit{SR} \\ \hline
  
  \multicolumn{2}{|c|}{$d=3$} & 0.977 & 0.619 & 1.000 & 0.977 & 0.619 & 1.000 & 0.975 & 0.618 & 0.998 & 0.971 & 0.617 & 0.982 \\ \hline
  
  \multicolumn{2}{|c|}{$d=10$} & 0.959 & 0.386 & 0.973 & 0.961 & 0.384 & 0.977 & 0.954 & 0.384 & 0.960 & 0.947 & 0.385 & 0.942 \\ \hline

  \end{tabular}}
  \caption{The simulation results of UN-CCDs on a set of synthetic datasets with varying noise level and 3 Gaussian clusters, demonstrating that the UN-CCD method is invariant to varying noise levels.} \label{tab:Noise_Level_Clustering_Apdx}
\end{table}

The UN-CCD method demonstrated remarkable robustness to background noise,
maintaining consistent performance across various noise levels.
Specifically, when $d=3$,
increasing the noise level ($l$) from $3\%$ to $20\%$ had virtually no impact on clustering quality.
For example, the \textit{ARI} values held steady above 0.97 when $l$ increases from $3\%$ to $20\%$;
meanwhile, the method successfully identified the correct number of clusters over $98\%$ of the time.
While the effect of noise is greater when $d=10$, 
UN-CCDs remain effective, with the \textit{SR} decreasing from 0.973 to 0.942.
Overall, these results demonstrate the UN-CCD's reliable performance in the presence of background noise across various dimensions.
These results confirm that UN-CCDs' geometric covering balls inherently isolate dense clusters from dispersed noise,
filtering out spatial contamination without the need for a separate outlier-removal preprocessing step.

\section{Real Data Examples}
\label{sec:Real-Data_clustering}

We assess the performance of three CCD-based clustering methods using some benchmark real and synthetic datasets in this section,
and compare them with some well-known, established clustering methods. 
Moreover, we want to explore if UN-CCDs outperform RK-CCDs after multiple enhancements.
The datasets, 
obtained from \textbf{UCI Machine Learning Repository} and \textbf{Clustering basic benchmark} \cite{uci_repository, ClusteringDatasets}, 
are generally more complex in the data structure, thus more challenging for clustering than the datasets used in Section \ref{sec:Simulation_Clustering}. 
To prevent features with larger scales from dominating,
we preprocess the datasets by normalizing them (except the synthetic datasets, including ``asymmetric", ``R15", and ``D31" datasets). 

\begin{figure}[htb]
    \centering
    \begin{subfigure}{0.31\textwidth}
        \includegraphics[width=\linewidth]{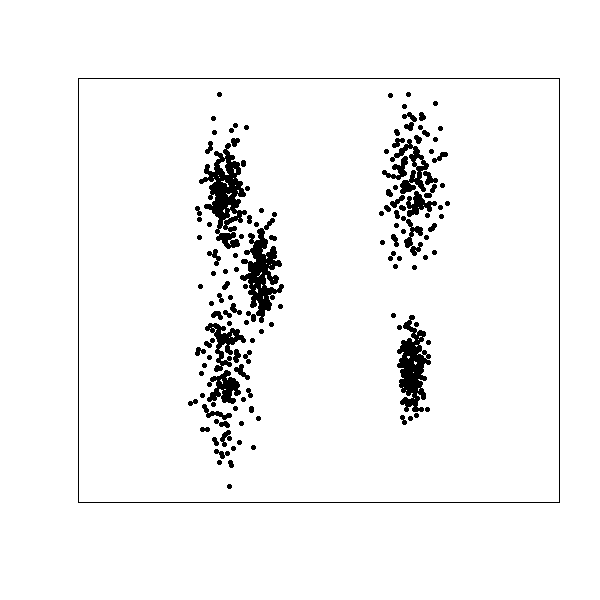}
        \caption{asymmetric}
        \label{fig:asymmetric_Apdx}
    \end{subfigure}
    \hfill
    \begin{subfigure}{0.31\textwidth}
        \includegraphics[width=\linewidth]{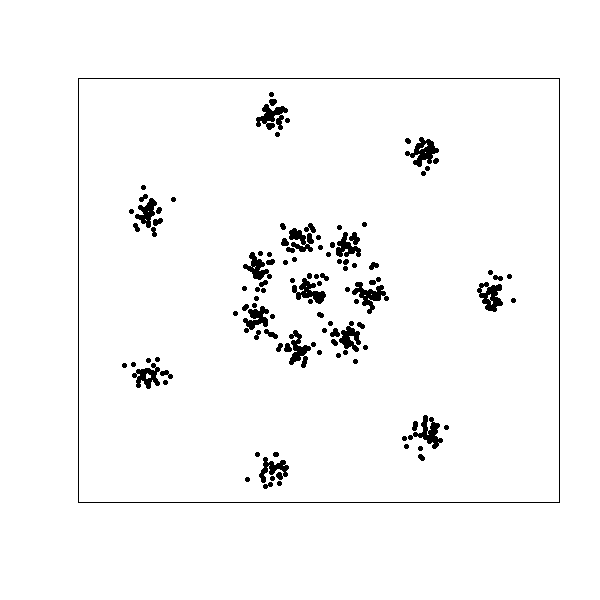}
        \caption{R15}
        \label{fig:R15_Apdx}
    \end{subfigure}
    \hfill
    \begin{subfigure}{0.31\textwidth}
        \includegraphics[width=\linewidth]{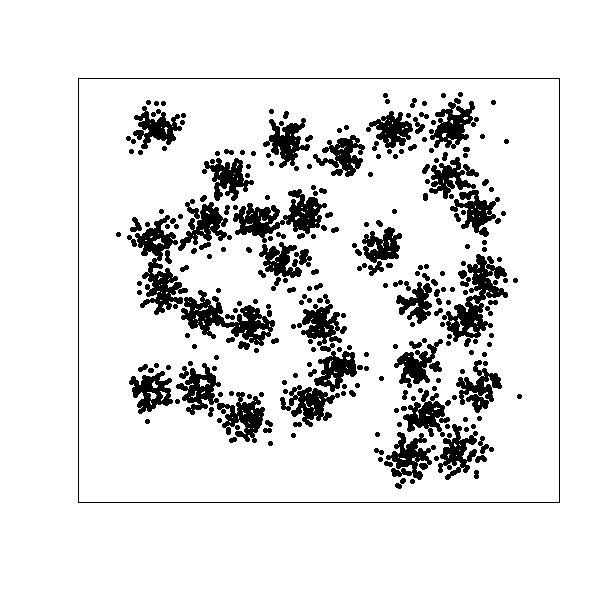}
        \caption{D31}
        \label{fig:D31_Apdx}
    \end{subfigure}
    \caption{The figures of asymmetric, R15, and D31 datasets}
    \label{fig:asymmetric_R15_D31_Apdx}
\end{figure}

The details and background of each dataset are summarized below.
Moreover, we present the ``asymmetric", ``R15", and ``D31" datasets in Figure \ref{fig:asymmetric_R15_D31_Apdx}. We are treating all datasets unlabeled, and compare the obtained clusters with the classes already available in the datasets. 


\noindent \textbf{Brief descriptions of each dataset.}
\begin{itemize} \label{tab:Real_Data_Des_clustering_Apdx}
  \item \textbf{iris}:  A dataset of iris plant, containing 3 types of 50 instances for each.
  \item \textbf{seeds}: The measurements of geometrical properties of kernels belonging to 3 different wheat varieties.
  \item \textbf{knowledge}: Measures the students' knowledge of Electrical DC Machines, categorized into 4 levels.
  \item \textbf{wholesale}: This dataset refers to clients of a wholesale distributor, it includes the clients' annual spending on diverse product categories, and are grouped to 3 regions.
  \item \textbf{asymmetric}: A synthetic 2-dimensional datasets with 5 elliptical Gaussian clusters of unbalanced cluster size and intensities, and the inter-cluster distances varies \cite{rezaei2020can}.
  \item \textbf{R15} and \textbf{D31}: Synthetic datasets consist of 15 or 31 similar Gaussian clusters \cite{veenman2002maximum}
\end{itemize}

\begin{table}[htb]
  \centering
  \begin{tabular}{|c|c|c|c|}
  \hline
  & $n$ & $d$ & \# of clusters \\ \hline
  iris      & 150  & 4  & 3 \\ \hline
  seeds     & 210  & 7  & 3 \\ \hline
  knowledge & 258  & 6  & 4 \\ \hline
  wholesale & 440  & 8  & 3 \\ \hline
  R15       & 600  & 2  & 15\\ \hline
  asymmetric& 1000 & 2  & 5 \\ \hline
  D31       & 3100 & 2  & 31\\ \hline
 \end{tabular}
 \caption{The size ($n$), dimensionality ($d$), and number of clusters of each real-life dataset.}\label{tab:Real_Data_clustering_Apdx}
 \end{table}


We compare the performance of CCDs with established clustering algorithms, \textbf{Density Based Spatial Clustering of Applications with Noise} (DBSCAN) \cite{ester1996density}, the \textbf{Minimal Spanning Tree} (MST) Method \cite{MST}, \textbf{Spectral Clustering} (Spectral) \cite{ng2001spectral}, \textbf{K-means++} \cite{arthur2006k}, and \textbf{Louvain Clustering Method} (Louvain) \cite{blondel2008fast}.

DBSCAN is a density-based clustering method proposed by Ester \textit{et al.} \cite{ester1996density},
often used to handle datasets with background noise or outliers.
It identifies clusters by finding ``seeds",
which are the points deep inside clusters that have a minimum number (denoted as \textbf{MinPts}) of neighbors within a given radius (denoted as \textbf{Eps}).
After that, it constructs clusters by finding all the points that are \textbf{density-reachable} from seeds.
Finally, the points not connected to any seeds are labeled as background noise or outliers.
\textbf{MinPts} and \textbf{Eps} are the two input parameters,
we follow a heuristic offered by Ester \textit{et al.},
which sets \textbf{MinPts} to 4,  
and find the value at the first ``elbow" of the sorted $4$-dist (the distance of a point to its $4^{th}$ nearest neighbor) plot, 
setting it to \textbf{Eps} \cite{ester1996density}.

The MST clustering method employs a graph-based approach \cite{MST}.
This method involves constructing a graph by connecting all data points with the minimum sum of edge weights, 
typically represented by the distance between two points.
Subsequently, edges with significantly larger weights than their adjacent edges' average weight are identified as "inconsistent" and removed.
This removal effectively breaks the MST into subtrees, 
each representing a distinct cluster.
After the trial-and-error process,
we set the threshold value for ``inconsistent" edges to 2,
which delivers the best overall performance compared to other thresholds.

Spectral clustering starts by constructing a similarity matrix of the given dataset \cite{ng2001spectral}.
Then, it derives the Laplacian matrix from the similarity matrix,
and the eigenvalues of the Laplacian matrix provide a low-dimensional embedding of the original data.
Finally, it applies a traditional method for clustering.
In our experiment,
we chose Gaussian similarity function to define the similarity
and $K$-means for clustering,
with the number of clusters $k$ as the input parameter.

$K$-means++ enhances the classic $K$-means clustering method \cite{arthur2006k}.
A key weakness of $K$-means is the sensitivity to the initially chosen cluster centroids.
$K$-means++ addresses this limitation by selecting initial centroids that are far apart,
leading to better and more stable clustering results and faster convergence.
Similar to spectral clustering,
we conduct $K$-means++ on each dataset with the true number of clusters as an input.

Louvain clustering method is a greedy algorithm finding communities in large networks by maximizing the ``modularity", which measures the quality of a partition\cite{blondel2008fast}.
The algorithm first optimizes modularity by iteratively moving individual nodes between communities, 
then aggregates these communities into new nodes and repeats the process until no further improvement is possible.
In this experiment, 
we first employ the standardized Gaussian kernel to construct a network for the given dataset, 
and then apply the Louvain clustering method with the default resolution parameter to find the communities (i.e., 1).

Consider the CCD-based methods,
$\alpha$ is set to $1\%$ for all the dataset for simplicity.
For KS-CCDs,
we choose the $\sqrt[d]{\delta}$ value that maximizes the average silhouette index in each dataset.

We choose the Adjusted Rand Index and Silhouette Index as external and internal validation measures.
Furthermore, we record the number of clusters detected by each method.
The performance of all methods is summarized in Table \ref{Real_Data_Result_Clustering_Apdx}.

\begin{table}[htb]
  \resizebox{\textwidth}{!}{\begin{tabular}{|c|c|c|c|c|c|c|c|c|c|c|c|c|c|c|c|c|c|c|c|c|c|}
  \hline
  \multirow{2}*{} & \multicolumn{3}{|c|}{iris ($k$=3)} & \multicolumn{3}{|c|}{seeds ($k$=3)} & \multicolumn{3}{|c|}{knowledge ($k$=4)} & \multicolumn{3}{|c|}{wholesale ($k$=3)}  & \multicolumn{3}{|c|}{asymmetric ($k$=5)} & \multicolumn{3}{|c|}{R15 ($k=15$)} & \multicolumn{3}{|c|}{D31 ($k=31$)} \\ \cline{2-22}
  & \textit{ARI} & \textit{Sil} & $\hat{k}$ & \textit{ARI} & \textit{Sil} & $\hat{k}$ & \textit{ARI} & \textit{Sil} & $\hat{k}$ & \textit{ARI} & \textit{Sil} & $\hat{k}$ & \textit{ARI} & \textit{Sil} & $\hat{k}$ & \textit{ARI} & \textit{Sil} & $\hat{k}$ & \textit{ARI} & \textit{Sil} & $\hat{k}$ \\ \hline
  KS-CCDs & 0.561 & 0.578 & 2 & 0.497 & 0.468 & 2 & 0.005 & 0.254 & 2 & -0.013 & 0.765 & 2 & 0.691 & 0.643 & 4 & 0.989 & 0.753 & 15 & 0.948 & 0.574 & 31 \\ \hline
  RK-CCDs & 0.557 & 0.393 & 3 & 0.500 & 0.164 & 3 & 0.052 & 0.119 & 2 & -0.002 & 0.344 & 2 & 0.730 & 0.605 & 4 & 0.993 & 0.753 & 15 & 0.814 & 0.495 & 30 \\ \hline
  UN-CCDs & 0.599 & 0.445 & 3 & 0.509 & 0.467 & 2 & 0.091 & 0.158 & 2 &  0.002 & 0.113 & 3 & 0.736 & 0.606 & 4 & 0.975 & 0.742 & 15	& 0.799 & 0.474 & 31 \\ \hline
  DBSCAN  & 0.544 & 0.517 & 2 & 0.003 & 0.121 & 1 & 0.005 & 0.198 & 1 &  0.006 & 0.569 & 1 & 0.467 & 0.548 & 3 & 0.936 & 0.702 & 15 & 0.350 & 0.343 & 15 \\ \hline
  MST     & 0.530 & -0.145 & 6 & 0.000 & -0.364 & 4 & 0.003 & -0.089 & 3 & 0.004 & -0.491 & 13 & 0.785 & -0.160 & 12 & 0.536 & -0.071 & 65 & 0.511 & -0.199 & 276 \\ \hline
  K-means++& 0.590 & 0.459 & 3* & 0.773 & 0.404 & 3* & 0.202 & 0.169 & 4* & -0.007 & 0.446 & 3* & 0.882 & 0.645 & 5* & 0.894 & 0.678 & 15* & 0.912 & 0.547 & 31* \\ \hline
  Spectral& 0.555 & 0.475 & 3* & 0.679 & 0.349 & 3* & 0.189 & 0.120 & 4* & -0.012 & 0.446 & 3* & 0.975 & 0.623 & 5* & 0.899 & 0.611 & 15* & 0.778 & 0.332 & 31* \\ \hline
  Louvain & 0.630 & 0.436 & 3 & 0.722 & 0.394 & 3 & 0.149 & 0.155 & 5 & -0.004 & 0.243 & 6 & 0.805 & 0.479 & 8 & 0.604 & 0.644 & 10 & 0.371 & 0.446 & 9 \\ \hline
\end{tabular}}
 \caption{The performance of selected clustering algorithms on real and complex datasets. \textit{ARI}: Adjusted Rand Index; \textit{Sil}: Silhouette Index; $k$: number of clusters ($k$ is the input parameter for K-means++ and spectral clustering); $\hat{k}$: estimated number of clusters.}\label{Real_Data_Result_Clustering_Apdx}
\end{table}

\paragraph{``iris" datasets:} Except MST, all the methods obtain comparable \textit{ARIs} on the ``iris" dataset.
Precisely, Louvain and UN-CCDs attain the highest \textit{ARIs} of 0.630 and 0.599, respectively.
KS-CCDs and DBSCAN produce the best \textit{Sil} values.
RK-CCDs, UN-CCDs, and Louvain clustering method correctly identify the true number of clusters.
MST's underperformance is due to its sensitivity to the varying intensities of the iris clusters.

\paragraph{``seeds" datasets:}
In the ``seeds" dataset, 
$K$-means++, spectral clustering, and Louvain clustering method perform better than other methods when considering \textit{ARI}.
On the other hand, UN-CCDs and KS-CCDs achieve the highest \textit{Sil} values of 0.468 and 0.467, respectively,
indicating better-defined cluster structures.
Both RK-CCDs and Louvain clustering method successfully identify the correct number of clusters.
DBSCAN underperforms due to the difficulty to find a single \textbf{Eps} value that works globally.

\paragraph{``knowledge" dataset:}
In the ``knowledge" dataset,
The performance of all the methods is generally mediocre.
$K$-means++ and spectral clustering get the highest \textit{ARI} values,
0.202 and 0.189, respectively.
KS-CCDs and DBSCAN produce the best \textit{Sil} values.
MST is the only method that identified 3 clusters.

\paragraph{``wholesale" dataset:}
All the methods perform poorly on the ``wholesale" dataset,
Suggested by the \textit{ARIs}, which are around 0,
indicating the clustering results akin to random assignment.
Therefore, we consider \textit{Sil} for evaluation,
KS-CCDs and DBSCAN outperform other methods, achieving \textit{Sil} values of 0.765 and 0.569, respectively.
Despite having the second-lowest \textit{Sil} value (0.113), 
UN-CCDs is the only method that correctly identifies the number of clusters. 

\paragraph{``asymmetric" dataset:}
When the number of clusters is known,
$K$-means++ and spectral clustering exhibit the best performance on the ``asymmetric" dataset.
These methods achieve \textit{ARI} values of 0.882 and 0.975, respectively, 
and \textit{Sil} values of 0.645 and 0.623, respectively.
The CCD-based methods show slightly lower performance, 
with \textit{ARI} and \textit{Sil} values around 0.7 and 0.6, respectively.
However, these methods identify 4 clusters, 
the closest to the actual number.
None of the CCD methods identify all five clusters because the three leftmost clusters, 
which have long elliptical shapes (Fig. \ref{fig:asymmetric_Apdx}), 
are too close to be reliably distinguished.
Datasets with small inter-cluster proximity and irregular cluster shapes remain challenging.

\paragraph{``R15" dataset:}
The three CCD-based methods perform better than the other methods on the ``R15" dataset, 
they successfully capture 15 Gaussian clusters, 
achieving \textit{ARI} values close to 1 and \textit{Sil} values around 0.75.
DBSCAN, $K$-means++, and spectral clustering deliver slightly inferior performance, 
with \textit{ARI} values around 0.9 and \textit{Sil} values around 0.65.

\paragraph{``D31" dataset:}
Consider the ``D31" dataset,
KS-CCDs and $K$-means++ detect all 31 clusters, 
deliver the highest \textit{ARI} values of over 0.9,
outperforming other methods.
RK-CCDs and UN-CCDs also perform well, 
with \textit{ARI} values of 0.814 and 0.799, respectively. 
They detect 30 and 31 clusters, respectively, 
outperforming all other methods except KS-CCDs and $K$-means++.

\paragraph{Summary of the Experiment:}

Among the evaluated methods, 
KS-CCDs consistently achieved the highest \textit{Sil} scores and competitive \textit{ARI} values across most real-world datasets. 
However, this high performance relies on optimally tuning the density parameter $\delta$, 
which incurs a steep computational cost. 
In contrast, the parameter-free variants---RK-CCDs and UN-CCDs---delivered highly robust results, 
matching or slightly exceeding the accuracy of $K$-means++ and spectral clustering, 
even when the latter were explicitly provided with the true number of clusters. 
Notably, UN-CCDs demonstrated a performance advantage over RK-CCDs in these complex scenarios. 
Standard density-based and modularity-based approaches such as DBSCAN and Louvain, 
yield slightly weaker partitions compared to RK-CCDs. 
Finally, MST consistently ranked lowest, 
frequently producing negative \textit{ARI} and Silhouette metrics due to its severe sensitivity to background noise and varying cluster intensities.

\section{Summary and Conclusion}
In this paper, we propose UN-CCDs, a CCD-based clustering method that replaces the Ripley's $K$-based spatial randomness test with a nearest-neighbor-distance (NND) based Monte Carlo procedure. This modification addresses key limitations of existing CCD variants and extends their applicability to moderate-dimensional settings.

Through extensive Monte Carlo experiments and evaluation on benchmark datasets, we show that UN-CCDs achieve stable and competitive performance relative to existing CCD-based methods and several standard clustering approaches within the considered regimes. In particular, the method performs well in scenarios characterized by moderate dimensionality, moderate sample sizes, complex cluster geometry, and the presence of uniformly distributed background noise. The empirical results also show a clear advantage over RK-CCDs as dimensionality increases into the moderate range, while retaining the practical benefit of requiring little parameter tuning.

At the same time, the method involves important trade-offs that define its scope of applicability. The computational complexity, on the order of $O(n^3)$, limits its use to datasets of moderate size, and the reliance on distance-based structure implies reduced effectiveness in higher-dimensional regimes where such measures degrade. In addition, the method is primarily designed for handling background noise rather than general forms of structured outliers, and its performance may vary in settings with overlapping clusters, heterogeneous noise patterns, or low-intensity clusters in close proximity.

Overall, UN-CCDs provide a useful addition to the class of graph-based clustering methods, particularly for problems in which geometric structure and robustness to background noise are central considerations. Future work may focus on improving computational efficiency, extending the method to more challenging high-dimensional settings, and developing adaptive strategies for parameter selection.

\paragraph{Future Work:}
Promising directions for future work include:
\begin{itemize}
  \item[(1)] Extending UN-CCDs to overlapping-cluster settings through ideas related to fuzzy clustering;
  \item[(2)] applying CCD-based clustering in semi-supervised settings with partially labeled data;
  \item[(3)] developing adaptive neighborhood or radius selection strategies based on local density;
  \item[(4)] automating the choice of the significance level ($\alpha$) in the MC-SRT with NND;
  \item[(5)] reducing the computational burden of the method while preserving its geometric advantages; and
  \item[(6)] improving effectiveness in substantially higher-dimensional settings, potentially through alternative distance or similarity measures.
\end{itemize}

\section{Acknowledgements}
Most of the Monte Carlo simulations in this paper were completed in part with the computing resource provided by the Auburn University Easley Cluster.
The authors are grateful to Art{\"u}r Manukyan for sharing the codes of KS-CCDs and RK-CCDs.

\section*{Declaration of generative AI and AI-assisted technologies in the writing process}
During the preparation of this work, the author used ChatGPT (OpenAI) to assist with drafting, rephrasing, and refining portions of the text for clarity and conciseness. After using this tool, the author thoroughly reviewed and edited the content to ensure accuracy, coherence, and alignment with the scholarly intent of the work, and takes full responsibility for the content of the publication.

\clearpage
\bibliographystyle{plain}
\bibliography{References}
\end{document}